\documentclass{report}

\input xy 
\xyoption{all}

\author{Bayle Shanks}
\title{Speculation on graph computation architectures and computing via synchronization}

\usepackage{amsmath} % pkg required by EasyLatex
\usepackage{amsfonts} % pkg required by EasyLatex
\usepackage{amssymb} % pkg required by EasyLatex
\usepackage{psfrag} % pkg required by EasyLatex
\usepackage[dvips]{graphicx} % pkg required by EasyLatex
\begin{document}
\maketitle
 
Note: Updated versions of this document, document source files, and related documents may be available at 
 
http://purl.net/net/bshanks/work/papers/computingWithGraphsAndSynchronization2004/
 
I recommend that you check that site before reading this; perhaps someday there will be a rewritten, more concise version, or a pointer to a subsequent paper, which could save you time.

\tableofcontents

\part{Main report}
\chapter{Introduction}
\section*{ Introduction: computing with graphs }
For this project, I tried to take a step back and explore some ways that the brain might implement simple computations. Rather than being driven by biological constraints, this project was an attempt to think of some unexpected and/or unlikely architectures. 

The topic is computing with graphs. Given that computation is being realized via a structure of nodes connected by edges, what sort of mechanisms might be used to compute? More specifically, at each time step there is a graph representing the current state of the network. There is some set of "update rules" which, when given the state of the graph at time $t$, will tell you what its state will be at time $t+1$. See Appendix \ref{appDefn} for a more formal definition of graph computation.

My goal, then, is to explore what types of graphs, graph structures, and update rules might be suitable for computation. Clearly, there are an infinite variety of possible setups that would have Turing-equivalent computational power. It may still be valuable, however, to look at a few exemplars from this infinite set in order to strengthen one's intuition about what might be possible\footnote{Eventually it may be valuable to come up with some mathematical proofs which detail some of the structure of the class of ways that one might compute with graphs. I haven't gotten there yet, though.}. 

\section*{ Turing machines }
A Turing machine is a mathematical construct which gives some rigor to the question, "Can this machine compute?". 

We want to to find computing architectures which are as powerful as Turing machines\footnote{I did not rigorously prove that any of the architectures were Turing-equivalent. This might be a good thing to do sometime.}. I've found\footnote{I haven't proved them but I think they're correct} some simple criteria for when a graph computation architecture will be as powerful as Turing machines. It is necessary and sufficient that such an architecture be able to express\footnote{as a subgraph} a NAND gate, be able to express a COPY gate, and be able to compose them. See Appendix \ref{appTuring} for much more elaboration on Turing machines and on when a graph computation architecture is as powerful as a Turing machine.  

\section*{ General remarks on graph computation architectures }
See Appendix \ref{appDefn} for my formal definition of a "graph computation machine". 

\subsection*{ Locality }
We want to impose some sort of locality restriction on the update rule. There are many kinds of locality restrictions that we could impose.

For example, we could require that the change of state of a node be determined only using information about the state of the arcs connected to that node. A more permissive version of this would allow information about the state of neighboring nodes to be used. Similar update rules could be applied to the state of arcs.

More interesting notions of locality would allow access to other "local graphical structure". For instance, while recalculating the state of a node, we may allow ourselves to ask if that node is a member of any cycle. This information is not available merely by considering the state of neighboring nodes and the arcs between them, yet it is, in a sense, "local", because there could still be regions of the network about which the node has no information:

\begin{figure}[h]
\centering
\includegraphics{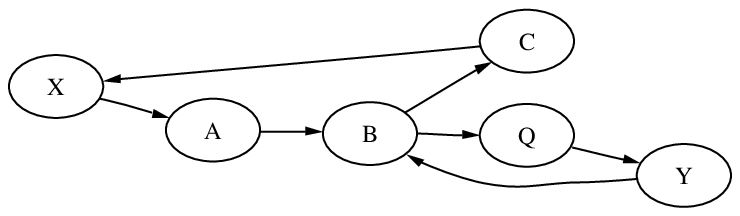}

\caption{In this example, the connectivity of Q and Y is irrelevant to the question of whether X is in a cycle. We are assuming that Q and Y are forbidden from ever connecting to X, A, or C.} 
\end{figure}

\subsection*{ Connection to cellular automata }
Graph computation, as I have defined it here, is a generalization of cellular automata. See Appendix \ref{appCell} for details.

Notably, there has already been some work on cellular automatons which can contain machines of Turing-equivalent power. For example, Conway's Game of Life can support Turing-complete machines.

\subsection*{ Fundamental graph structures }
Conventional neural networks focus on nodes as the primary computational elements. But what other structures are important in graphs? 

\begin{itemize}
\item[*] Nodes
\item[*] Edges (also called arcs)
\item[*] Cycles
\item[*] Paths
\item[*] Triangles
\item[*] Cuts, flows
\item[*] Cliques
\item[*] Partitions
\end{itemize}

Maybe we should try to think of ways that computation may be accomplished using some of these other choices as the basic elements.

\chapter{Arc computation}
This chapter is about computation which uses arcs, not nodes, as the basic computational elements. Conventional neural networks may be visualized as pulses of activation traveling down a line from one node to another. The pulses arrive at nodes, which perform some simple computation, and then perhaps emit a pulse of their own, which travel to other nodes.

Instead of visualizing pulse of activation traveling down lines, visualize the lines themselves flickering on and off. In an ideal arc computation network,  none
of the computational "action" is in the individual impulses, and all of the action is in the changing structure of the connections between the nodes.

\section*{ Gates }
What are the analogs of "logic gates" in arc computation?

The analog of binary values is the presence or absence of an edge in the graph.

Gates can often be formulated as subgraphs. Define some potential arcs in the subgraph as "input ports" and some potential arcs as "output ports". Assume that we have chosen a fixed update rule. If, for this update rule, the effect of the subgraph's input ports on its output ports is the same no matter what bigger graph it is embedded in\footnote{out of the allowed graphs for the chosen graph computation machine. That is to say; you are allowed to "disallow" certain graph structures so that they don't break your gates!}, then we can use the subgraph like a subroutine. A logic gate is just a very simple subroutine.

\begin{figure}[h]
\centering
\includegraphics{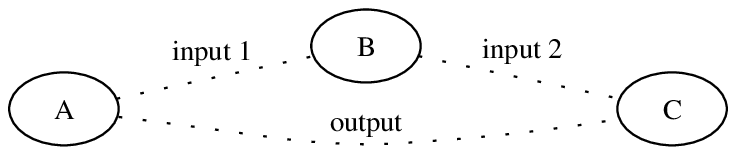}
\caption{Triangular logic gate schematic for arcs. The dotted lines represent potential arcs.}
\label{gate1}
\end{figure}

For example, for some update rules the subgraph in Fig. \ref{gate1} functions as a NAND gate.

For the structure in Fig. \ref{gate1} to work like a NAND gate, it must always satisfy the truth table in Fig. \ref{ttTriNAND}.

 \begin{figure}[h]
\centering
\begin{tabular}{|l|l|l|}
\hline && 
\\ Arc between A and B & Arc between B and C & Result at time $t+1$  
\\ at time $t$ & at time $t$  &
\\ \hline && 
\\   Absent & Absent & There is an arc between A and C
\\ Absent & Present & There is an arc between A and C
\\ Present & Absent & There is an arc between A and C
\\ Present & Present & There is no arc between A and C
\\ \hline
\end{tabular}
\caption{NAND gate truth table for triangular logic gates made of arcs}
\label{ttTriNAND}
\end{figure}

\begin{figure}[h]
\centering
\includegraphics{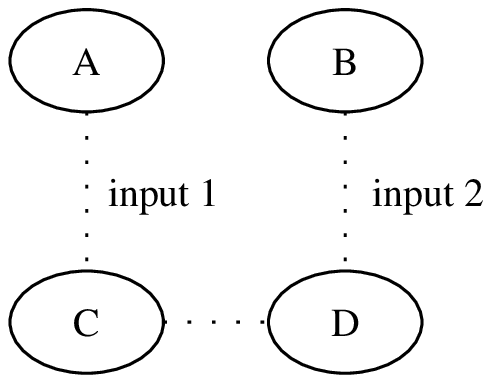}
\caption{A logic gate schematic. CD is the output. The dotted lines represent potential arcs.}
\label{4nodeSchm}
\end{figure}

Triangles form a natural structure for gates because they have three sides. However, implementing triangle gates in certain physical systems may give us grief because two of the input sides share a node. So we may want to use a four-node "square" structure instead, as in Fig. \ref{4nodeSchm}.

And for the structure in the last figure to work like a NAND gate, it must always satisfy the truth table in Fig. \ref{ttSqNAND}.

\begin{figure}
\begin{tabular}{|l|l|l|}
\hline && 
\\ Arc between A and C & Arc between B and D & Result at time $t+1$  
\\ at time $t$ & at time $t$  &
\\ \hline && 
\\   Absent & Absent & There is an arc between C and D
\\ Absent & Present & There is an arc between C and D
\\ Present & Absent & There is an arc between C and D
\\ Present & Present & There is no arc between C and D
\\ \hline
\end{tabular}
\caption{NAND gate truth table for "square" logic gates made of arcs}
\label{ttSqNAND}
\end{figure}

\section*{ Neural interpretations of arcs }
\subsection*{ Arcs as synapses; plasticity }
If arcs are synapses, then arc computation provides a theoretical model for computing with \emph{plasticity} instead of with action potentials. Perhaps in some parts of the brain, the "fundamental unit of data" is synapse strength, not activation\footnote{Of course, for plasticity to be fast enough, we would probably want to model not just long-term plasticity but also short-term plasticity such as faciliation and depression. But then again, maybe some brain areas just compute slowly.}.

\section*{ Arcs as synchronization }
In the brain, we might interpret nodes as neurons, and arcs as representing \emph{synchrony} between neurons within a time step. This will allow us to model\footnote{Synchronization is a complex set of phenomena which cannot be completely modeled without tools like differential equations and nonlinear dynamics. We will abstract away from the actual physics of synchronization. Instead, we're looking for a "qualitative physics" of synchronization. "Qualitative physics" is a  cognitive science term which basically means abstracting most of the numbers away and only modeling the state transition diagram between different regimes of the system.} synchronization-based neural computation with graphs.

These "implementation details" are separate from the question of what kinds of Turing-complete graph computation machines exist, so I've put them in a separate chapter. Often, a qualitative physics is not itself an update rule, but is rather an explanation for why the update rule is what it is, or how the update rule might be implemented in a physical system; that's why they are "implementation details".

For other general notes on arc computation, see Appendix \ref{appArc}.

\chapter{Computing with synchronization}

This chapter considers a special case of arc computation, that is, when the arcs represent synchronization between two nodes. It covers what synchronization might mean, and points to appendices which give simple models of neural synchronization, and logic gates implemented in those models.

\section*{ Definition of exact synchrony }
Consider a population of neurons evolving over some short period of time $\Delta t$. What sort of notions of "synchronization" might help us compute?

For any two neurons $a$ and $b$, if the output $a$ and $b$ over $\Delta t$ are identical, we say that $a$ and $b$ are \emph{exactly synchronized}. 

Sometimes the only output that we are concerned about is spike times. Let $A$ be the set of spike times of $a$ during $\Delta t$, and let $B$ be the set of spike times of $b$ during $\Delta t$. 

Exact synchronization is reflexive, symmetric, and transitive. It is an \emph{equivalence relation}, and at any given time\footnote{with an associated time window}, the relation of exact synchronization partitions the network into equivalence classes\footnote{i.e. groups of neurons which are doing exactly the same thing during the time window}. 

If we visualized exact synchronization between nodes by drawing an arc between nodes when they were exactly synchronized, then the graph would consist entirely of cliques\footnote{a clique is a subset of fully-connected nodes}. For example, if we have nodes $a, b, c, d, e$, and $a,b,c$ and $d,e$ were exactly synchronized near some time $t$, then we might represent the synchronization state of the network with this graph:

 \begin{figure}[h]
\centering
\includegraphics{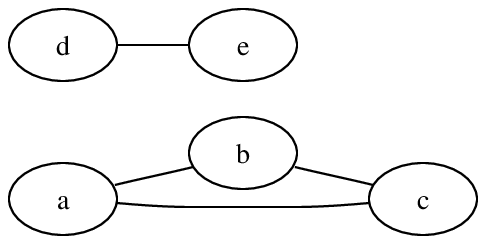}
\caption{A graphical representation of the state of some network near time $t$, where arcs denote exact synchrony at time $t$}
\end{figure}

See Appendix \ref{appMoreSyncDefn} for some other notions of synchrony that might be useful, and a few notes on their mathematical structure.

Sometimes I'll use the term "sync arc" to mean an edge denoting exact synchronization between two neurons. These "sync arcs" are different from the underlying "connective edges" of the network. Sync arcs represent the presence of synchronization at some time step, whereas connective edges represent a synaptic connection. I'll usually draw sync arcs as dotted lines.

\subsection*{ Logic gates for exact-synchrony }
How might we implement "logic gates" with exact-synchrony based computing? See Appendix \ref{synchGates1} for implementations of the fundamental boolean gates, and Appendix \ref{nonboolSynchGates} for implementations some other, non-boolean gates. Those sections also contain 3 distinct "qualitative physics" models of synchronization.

\section*{ Synchrony in nonlinear oscillators: a dynamical view }
In Appendix \ref{appDynamic}, you'll find a definition of another kind of synchrony, \emph{dynamic synchronization}, which is applicable to forced nonlinear oscillators. A common phenomenon in such systems, the "Devil's staircase", is introduced. Also, a brief definition of "circle maps" is given; circle maps are simple, well-studied discrete dynamical systems which exhibit mode-locking and the Devil's staircase, and which are special cases of integrate-and-fire neurons.

\subsection*{ Logic gates for dynamic synchrony }
\begin{figure}[h]
\centering
\includegraphics{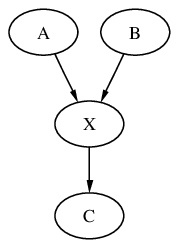}
\caption{What a logic gate might look like using dynamic synchronization. Each node above represents a nonlinear oscillator, and the arrows represent forcing. When A and X are in dynamic synchronization, then we say that "input 1 is TRUE". When B and X are in dynamic synchronization, we say that "input 2 is TRUE".  When X and C are in dynamic synchronization, we say that "the output is TRUE".}
\label{nonlinGate}
\end{figure}

Perhaps we could use the presence or absence of dynamic synchronization to represent binary data. In this framework, a logic gate might be represented as a nonlinear oscillator receiving input from two forcing oscillators. The oscillator would itself be forcing a third oscillator. The synchronization or lack of synchronization between these oscillators would represent binary inputs and output. Fig. \ref{nonlinGate} shows how such a logic gate might look.

\section*{ More complicated synchronization schemes }
Appendix \ref{appMisc} has a few other notes on other ways to model synchronization.

\chapter{Computing with cycles}
Another fundamental unit that we could compute with is cycles. There are at least three ways that we could look at cycles:

\subsection*{ Am I part of a cycle? }
Have a notion of locality in which a node can "know" not just its neighbors, but also if it is involved in a cycle, and perhaps which of its arcs are involved in a cycle. Binary logic would be represented as whether a node is part of any cycle, or whether it is not. Each node represents one bit; 0 means not a part of any cycle, 1 means part of a cycle.

\subsection*{ Is this cycle active? }
Another way to think about things would be to look at all potential cycles in the graph. Binary logic would be represented as whether each potential cycle was an actual cycle. Each potential cycle is one bit; the bit is 0 if the cycle is not active, and 1 if it is. 

\subsection*{ Is he my brother? }
A third way would be to have a notion of locality in which a node $a$ can know, for any other node $b$, if $a$ and $b$ are both part of some cycle. Note that each node has the potential to be members of many cycles.

\section*{ Cycle gates }
Fig. \ref{cycleGate1} shows what a gate might look at in an Is-This-Cycle-Active system:
\begin{figure}[h]
\centering
\includegraphics{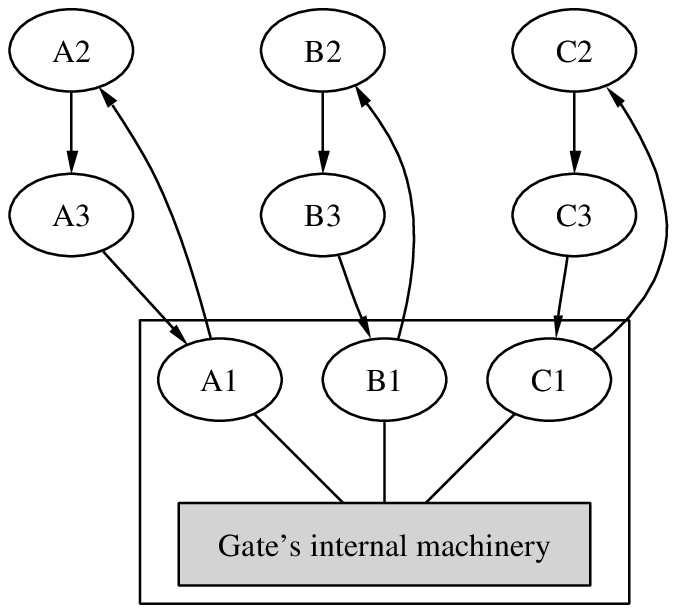}

\caption{What a two-input, one-output gate might look like. An intact cycle represents 1, a broken cycle represents 0. The inputs are the A cycle and the B cycle; the output is the C cycle. In this example, the A, B, and C cycles are all intact, so the input is TRUE, TRUE, and the output is TRUE}
\label{cycleGate1}
\end{figure}

For perspective, Fig. \ref{cycleGate2} shows what the same gate might look like if it were receiving FALSE, FALSE, and sending TRUE:
\begin{figure}[h]
\centering
\includegraphics{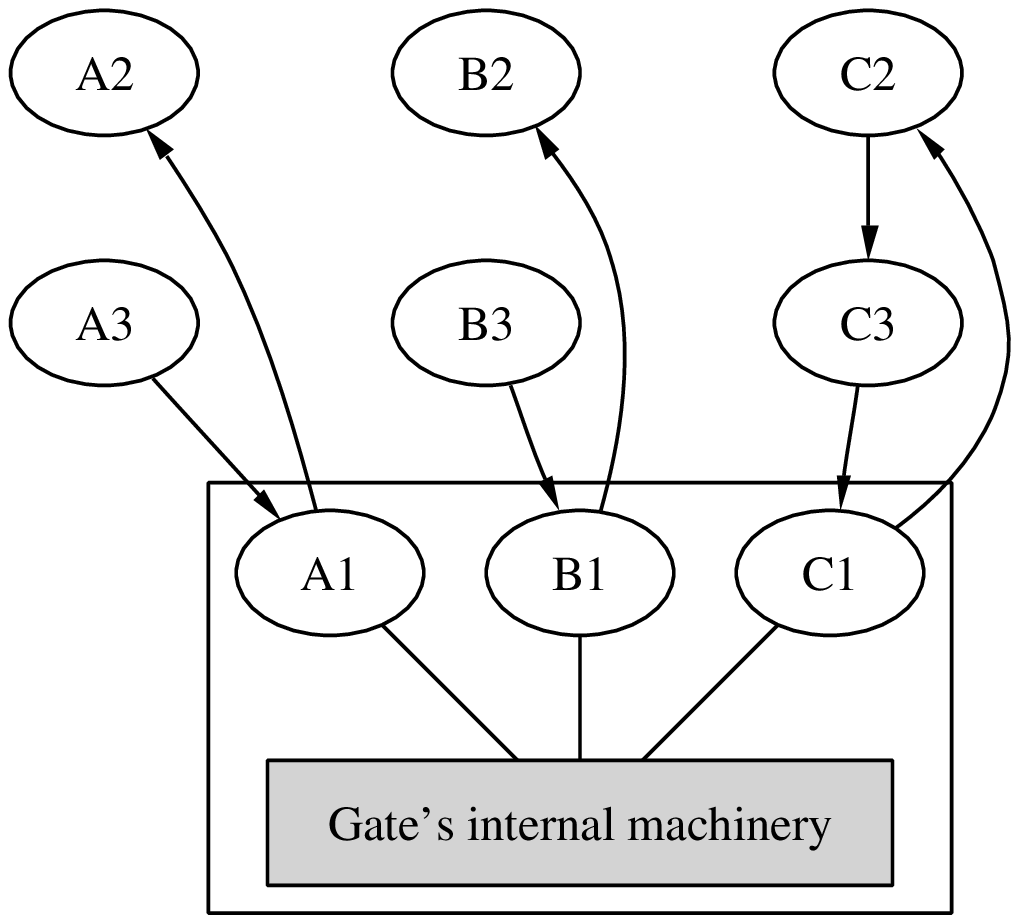}
\caption{What a two-input, one-output gate might look like. An intact cycle represents 1, a broken cycle represents 0. The inputs are the A cycle and the B cycle; the output is the C cycle. In this example, the A and B cycles are "broken", and C cycle is intact, so the input is FALSE, FALSE, and the output is TRUE}
\label{cycleGate2}
\end{figure}

\subsection*{ Implementation of a NAND gate }
See Appendix \ref{appCycleNAND} for a toy physics/update rule for doing cycle computation, and an implementation of a NAND gate within this framework.

\newpage

\chapter{References}

\subsection*{ Boolean functions \& Boolean circuits }
Vaughan Pratt's CS353 notes, http://boole.stanford.edu/cs353/handouts/book3.pdf

Ingo Wegner, "The Complexity of Boolean Functions, http://eccc.uni-trier.de/eccc-local/ECCC-Books/wegener\_book\_readme.html

\subsection*{ The Devil's Staircase and Synchronization in Dynamical Systems }
S. Coombes and P. C. Bressloff. Mode-locking and Arnold tongues in integrate-and-fire neural oscillators. http://www.lboro.ac.uk/departments/ma/preprints/papers99/99-9.pdf

Rodrigo Laje and Gabriel B. Mindlin. Highly Structured Duets in the Song of the South American Hornero. Phys. Rev. Lett. 91, 258104 (2003). http://www.santafe.edu/sfi/education/international/intlfellows/intlfal02/fellows/files/mindlin.pdf

Eric W. Weisstein. "Circle Map." From MathWorld--A Wolfram Web Resource. http://mathworld.wolfram.com/CircleMap.html

\subsection*{ Other exotic realizations of Turing machines }
\subsubsection*{ Chip Firing Games \& sandpiles }
\emph{Chip Firing Games are graph computation machines which work on directed graphs. They are Turing complete. To quote Goles '01 below, "$\ldots$ a number of chips is stored at each vertex, and if one vertex has at least as many chips as outgoing edges, then one chip is transferred from this vertex along each of these edges."}

\emph{They are of interest as generalizations of "sandpile models", which are studied in catastrophe theory (i.e. when the sand is piled too high and it falls, it's a catastrophe!). Also studied as models of aspects of parallel computation and peer-to-peer networks, and for fun.}

\emph{Note; I haven't looked too closely at most of the items in this section yet}

Anderson, Richard, Laszlo Lov?sz, Peter Shor, Joel Spencer, Eva Tardos, and Shmuel Winograd, ``Disks, Balls, and Walls: Analysis of a Combinatorial Game, `` American Mathematical Monthly, volume 96, June-July 1989, pages 481-493.

Bitar, Javier, Eric Goles ``Parallel chip firing games on graphs'', Theoretical Computer Science, 92, (1992), pp.291-300. 

Bjorner, Anders, Laszlo Lov?sz, and Peter Shor, ``Chip-firing games on graphs'', Europ. J. Combinatorics, 12 (1991), pp. 283-291. \emph{This paper introduced chip firing games.}

Angela R. Kerns. Sandpiles in Graphs. http://www.cs.wvu.edu/~angela/cs418a/cs418a.html

Ericksson, Kimo, ``Chip-firing games on mutating graphs'', SIAM J. Discrete Math., vol. 9, no. 1, February 1996, pp 118-128.

Goles, E. \& Latapy, M. Complexity of grain-falling models. 
\\http://www.complexity.org.au/ci/draft/draft/latapy01/latapy01s.ps

Goles, Eric, and Marcos A. Kiwi, ``Games on line graphs and sand piles'', Theoretical Computer Science, 115 (1993), pp. 321-349.

Eric Goles Ch., Maurice Margenstern: Universality of the Chip-Firing Game. Theor. Comput. Sci. 172(1-2): 121-134 (1997). \emph{Apparently this article proved that chip firing games are Turing-complete.}

Tardos, Gabor, ``Polynomial bound for a chip firing game on graphs,'' SIAM J. Discrete Math., vol. 1, no 3, August 1988, pp. 397-398.
Tomas Feder. Disks on a tree: analysis of a combinatorial game. (with C. Subi) submitted SIAM J. Discrete Math. http://theory.stanford.edu/~tomas/comb.ps

\newpage

\part{Appendices}
\appendix

\chapter{Definitions} \label{appDefn}

A \emph{graph} $G$ is a tuple $(N,E,nodelabels,edgelabels)$, where $N$ is a set, $E$ is a set of ordered pairs of $N$, $nodelabels$ is a function whose domain\footnote{The codomains of $nodelabels$ and $edgelabels$ may vary; in the most common case, the nodes and edges are labeled by integers, in which case the codomain of both these functions is $\mathbb{Z}$} contains $N$, and $edgelabels$ is a function whose domain contains $E$. 
\\
\\A \emph{graph update rule} is a function which takes graphs to graphs.
\\
\\A \emph{graph type} is a tuple $(allowedGraphs, nodeLabelSet, edgeLabelSet)$, where $allowedGraphs$ is a set of graphs, $nodeLabelSet$ is a set, and $edgeLabelSet$ is a set.
\\
\\A \emph{graph computation machine} is a tuple $(graphType, updateRule, inputFn, outputFn)$, where $graphType$ is a graph type, $updateRule$ is a function from $allowedGraphs$ to $allowedGraphs$, $inputFn: \mathbb{Z} \to allowedGraphs$, and $outputFn: allowedGraphs \to \mathbb{Z} \cup NOT\_HALTED\_YET$.
\\
\\If C is a graph computation machine and $I\in\mathbb{Z}$, then the \emph{computation time of C on input I} is 

\begin{align*}
(\mu t\in\mathbb{N})\textrm{ s.t. } \qquad outputFn(updateRule^t(inputFn(I))) \neq NOT\_HALTED\_YET
\end{align*}
\\
\\Let $T =$ the computation time of C on input I. 
The \emph{output of C when given input I} is 
NEVER\_HALTS if 
$T = \infty$, or 
$outputFn(updateRule^T(inputFn(I)))$ otherwise.
\\
\\Here is how to use these constructs. To compute with a graph computation machine, you give the $inputFn$ an input, and you get back a graph (which represents a graph computation machine initialized to run on that input). You iterate the update function on that graph until $outputFn$ tells you that the machine has halted. Then, $outputFn$ tells you the value of the output.
\\
\\Note that the update function may change the nodes and edges themselves, not just the labels on them.
\\
\\Now we define node-based and arc-based graph computation machine.
\\
\\Let $M$ be a graph computation machine. Let $M = (graphType, updateRule, inputFn, outputFn)$. If $updateRule$ never changes the labels of edges, and never adds or deletes edges or nodes\footnote{If we want to allow node deletion, we can usually simulate this by having a certain node label, such as 0 or 1, correspond to "deleted node"}, then we say that $M$ is \emph{node-based}. If $updateRule$ never changes the labels of nodes, and never adds or deletes nodes or edges\footnote{Again, if we want to delete edges, we can do pick update rules for which an edge label of "0" corresponds to an inactive edge, or do something else like that}, then we say that $M$ is \emph{arc-based}.

\chapter{Comparison with cellular automata} \label{appCell}
A cellular automata is a system with a grid of "cells", along with an update rule that tells you, given the state of each cell at time $t$, what the state will be at time $t+1$. Usually the state of each cell at time $t+1$ is only dependent on its current state and the state of its neighbors at time $t$.

The differences between cellular automata and graph computation machines are:

\begin{itemize}
\item[*] No tiling: The cells in a cellular automata are usually regularly tiled polygons across a (hyper)plane. The nodes in a graph do not have to have this structure. In particular, nodes in a network may have different numbers of "neighbors".
\item[*] More permissive notions of locality: The definition of "locality" in cellular automata refers to the neighborhood of each cell. In graphs, we might define locality differently, for instance, "any node B which share a cycle with node X is local to X".
\item[*] Dynamic topology: A graph may compute by changing its arcs, which may change the "local neighborhood" of graph elements dynamically. Cells in cellular automata have a fixed neighborhood for the during of the computation.
\item[*] Information in the arcs: A graph might attach labels to its arcs, whereas cellular automata usually do not store any information in the spaces between cells.
\end{itemize}

So, cellular automata are a special case of graph computation. For example, the set of 2D cellular automata is equivalent to the set of graph computation machines whose nodes are connected in a 2D grid (4 edges per node), whose edges have no labels, whose update rules only alter node labels, and whose update rules determine each node's new label by looking only at the labels of adjacent nodes.

\begin{figure}[h]
\centering
\input{easyLatexGraph7568.tex}
\includegraphics{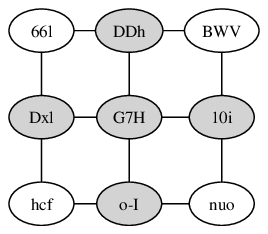}
\caption{Schema of a 9-cell 2D cellular automata when represented as a graph. Note that there are no edge labels. The filled in circles denote the local neighborhood of node $n_{2,2}$.}
\end{figure}

\chapter{More on Arc-based Computation} \label{appArc}

It is true that one could model the behavior of such a network as a standard "node-based" network computation. But this doesn't marginalize their importance. Node-based networks can be Turing-complete, and hence they can model any other known computational architecture, too. 

There is no "natural" way to convert every arc-based graph computation machine to a "dual" node-based graph computation machine\footnote{Because there is no natural dual for graphs. There are formal mathematical definitions of "naturality", such as category-theoretic naturality, which might be applied to prove that there is no natural dual; I haven't seen the proof. But see Circular Chromatic Number and Circular Flow Number of Graphs, Xuding Zhu, http://www.math.ntu.edu.tw/~gjchang/conference/2002-07-to-12-graph-workshop/2002-11-06-zhu-abs.doc}. For example, you would want to represent nodes in one graph by edges in the other, but how would you represent a node without any edges in the dual graph? Certainly there are information-preserving transformations, but there is no "natural" dual.

\subsection*{ Aiming for simplicity }
Since it would be nice to have a really elegant scheme, I made the following wish list:

\begin{itemize}
\item[*] Concise update rule
\item[*] The update rule obeys some notion of locality
\item[*] Uses undirected graphs
\item[*] No node labels
\item[*] Uses simple edge labels (i.e. single numbers)
\item[*] Update rule is well-defined and produces meaningful results for almost any network structure
\end{itemize}

Most of the update rules that I've found don't meet all of these criteria. And that's okay too.

\chapter{Some other neat types of synchrony} \label{appMoreSyncDefn}

We say that $a$ and $b$ are \emph{partially spike-synchronized} if either $A \subseteq B$ or $B \subseteq A$. Partial spike-synchronization is reflexive and symmetric but not transitive. A more useful notion might be to consider the full structure of which neurons' sets of spikes were subsets of which others. This structure is an example of a \emph{ordered set}, specifically, the ordered set induced by the subset relation on the sets of spike times. 

We might visualize the partial synchronization state of the network using a directed graph, with a directed arc $a \to b$ when $B \subseteq A$, and an undirected arc $a \leftrightarrow b$ when $a$ is in exact synchrony with $b$ ($A = B$)). Note that, in such a network, we could say that $a$ and $b$ are \emph{partially spike-synchronized} whenever there is \emph{any} arc between them, whether directed or undirected.

For example, the following graph might denote the partial spike-synchronization state of some network at a time $t$:

 \begin{figure}[h]
\centering
\includegraphics{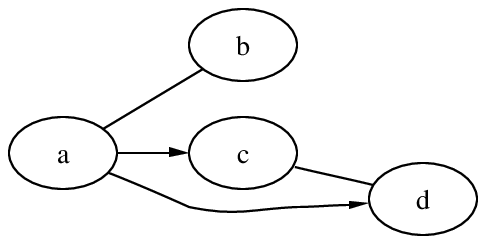}

\caption{A graphical representation of the partial spike-synchronization state of some network near time $t$. This is what we would draw if the spike times for unit $a$ were \{0, 5, 10, 15, 20\}, the times for $b$ were \{5, 15\}, the spike times for $c$ were \{0,10\}, and the spike times for $d$ were \{0,10\}.}
\end{figure}

Define the \emph{correlation} of the spike times of $a$ and $b$ as $\frac{|A \cap B|}{|A \cup B|}$. Note that we could have either partial spike-synchronization or high correlation without the other one.

\chapter{Nonlinear dynamical systems and synchronization} \label{appDynamic}

Consider a population of neurons evolving over some period of time $\Delta t$.

For any two neurons $a$ and $b$, say that we have some notion of a "coincidence". This might be whenever both neurons spike nearly instantaineously, or it might be whenever both neurons reach the peak of an oscillation at about the same time. 

Suppose that $a$ and $b$ are each exhibiting quasi-periodic behavior. Let $w_a$ and $w_b$ be the frequencies of $a$ and $b$. Let $p$ be the number of cycles of $a$ between consecutive coincidences, and let $q$ be the number of cycles of $b$ between consecutive coincidences. Define $r_{ab} = \frac{p}{q}$. Note that if $a$ and $b$ are periodic, then $r_{ab} = \frac{w_a}{w_b}$.

If $r_{ab}$ resists small perturbations of $w_a$ or if it resists small perturbations of $w_b$, then we say that $a$ and $b$ are in \emph{dynamic synchronization}\footnote{I believe that this is typically called "mode-locking", but I'm not totally sure, so to be cautious I'm calling it something else.}.

This sort of synchronization (amongst others) is studied in nonlinear dynamics\footnote{$r_{ab}$ is called a "rotation number"}. There are some typical phenomena which often turn up in nonlinear driven oscillators. One of them is called "the Devil's staircase". If you plot the period of one oscillator on one axis and $r_{ab}$ on the other one, you often get a plot that looks like Fig. \ref{devils}

\begin{figure}[h]
\centering
For copyright reasons, this figure omitted from free archive. Please refer to figure 2(b) at http://www.santafe.edu/sfi/education/international/intlfellows/intlfal02/fellows/files/mindlin.pdf .
\caption{The Devil's Staircase. Figure copied from Laje and Mindlin 2003.}
\label{devils}
\end{figure}

To quote\footnote{Laje and Mindlin, 2003}, "This organization is typical of nonlinear oscillators. Nonlinear and linear oscillators react under periodic forcing in a qualitatively different way. Linear oscillators end up following the forcing, while nonlinear ones display a wide variety of subharmonic behaviors, depending on the amplitude and frequency of the forcing. If the frequency of the forcing is similar to the natural frequency of the forced oscillator, they will lock in a one-one regime for a wide range of pa
rameter values. However, if the frequency of the forcing is larger than the natural frequency of the forced oscillator, other behaviors can be found. In the case the forcing amplitude is not too large, periodic and quasiperiodic motions are possible." 

The plateaus in this graph represent regions where small perturbations of the forcing frequency cause the forced oscillator to adjust its frequency so as to maintain the frequency ratio. In other words, the plateaus represent what we have here termed "dynamic synchronization".

The main text sketches how we might use dynamic synchronization to represent binary information, and compute using gates made up of linked nonlinear oscillators (although, for this system, I haven't actually constructed any gates).

\subsection*{ Circle maps }
One model of neurons which exhibits can exhibit mode-locking synchronization is integrate-and-fire neurons\footnote{S. Coombes and P. C. Bressloff. Mode-locking and Arnold tongues in integrate-and-fire neural oscillators. http://www.lboro.ac.uk/departments/ma/preprints/papers99/99-9.pdf}. In some regimes, the behavior of integrate-and-fire neurons can be described by "circle maps", a well-studied discrete dynamical model:

\begin{align*}
\theta_{t+t} = \theta_t + \Omega - \frac{K}{2 \pi} sin (2 \pi \theta_t)
\end{align*}

The parameter $\Omega$ can be interpreted as an externally applied frequency, and the parameter $K$ can be interpreted as a strength of nonlinearity. Circle maps display mode-locking and the Devil's staircase phenomenon.

I did not have a chance to delve deeper into this area, but I note it as an interesting area of future study. Can we construct logic gates (and therefore Turing machines) out of a network of interacting circle maps? I bet the answer will be "yes".

\chapter{Implementating logic gates with synchrony} \label{synchGates1}

\section*{ Modeling exact synchrony }
We'll use the following simple model of exact synchrony in a neural network. Assume that no two nodes are synchronized unless it is explicitly stated that they are\footnote{That is, nodes never just "happen" to synchronize; if there's not a reason to synchronize, we will assume that they are not synchronized. If you are worried about that, you could assume that each neuron has a slightly different "natural frequency" that it likes to fire at.} \footnote{Note that two nodes which are both silent are exactly synchronized by our definition}. 

Assume that two nodes which get synchronized input become synchronized.\footnote{Except for the "input nodes" on different ports of each gate, which are assumed to have different enough properties from each other that they are not synchronized unless we say they are. Specifically, if a gate has two input ports, $AB$ and $CD$, and receives an identical synchronized signal on each port, then we'll assume that $a$ is synch'd with $b$ and $c$ with $d$, but that nevertheless $a$ and $c$ are not sync'd, and either are $b$ and $d$. One might implement this by hhaving $a$ and $b$ fire 2 ms after they receive an input spike, whereas $c$ and $d$ fire 50 ms after they receive and input spike. This is to prevent a logic gate from malfunctioning if it gets two separate inputs from two separate places which are accidentally synchronized.} 

If a neuron gets multiple inputs, we will define what it's "total input" is. If a neuron $x$ gets as input sets $A, B, C\ldots $ of spike times as input, then the "total input" to which the above rules apply is $A \cup B \cup C\ldots $.

In summary, two nodes emit spikes at the same set of times if and only if they receive spikes at the same set of times\footnote{Clearly we aren't dealing with noise yet! For now, we're sticking to simple abstractions. Note, though, that this abstraction naturally extends to a notion of spike trains which are CLOSE to identical; so noise could probably be handled}.

There are also special "threshold units"\footnote{It would be elegant if we could express our model without putting "information into the nodes", and indeed we can. instead of threshold units, we could have threshold arcs; every arc would be a threshold arc, and the arc label is the threshold. Most arcs have an arc label of "1", which essentially means that the threshold is always met. To deal with the inhibitory neurons that we'll introduce later, we will say that a negative arc is an inhibitory arc. Now all of our nodes are identical, and the information is only in the arcs.}, each with a threshold $n$, which are silent unless they receive at least $n$ synchronous inputs on a single timestep. When threshold units do fire, they are not in sync with any other units (the same restrictions are on them as on input units; see below). Assume that no units are threshold units unless it is stated otherwise.  

Further, gates have "input" nodes which receive input external to the subgraph. So, we'll make the following assumptions about those inputs. 

\begin{itemize}
\item[*] Assume that no input is silent, i.e. every input is a nonempty set. 
\item[*] assume that the SET UNION of ANY proper subset of the input spike time sets does NOT match the set union of other proper subset of the inputs, except for set unions which must be identical because of stated synchronizations\footnote{This is pushing things a little, but we need it to ensure that we won't accidentally synchronize while computing with signals that are supposed to be different. With judicious set up of the instrinsic properties of the neurons generating the input signals, though, one can probably avoid accidental synchronization easily. Note that the current model puts no direct constraints on the I/O chacteristics of single neurons (we're only concerned about whether they match each other or not); this leaves a lot of freedom for different intrinsic properties.}. 
\end{itemize}

\section*{ Some boolean logic gates }
We will model boolean logic where the presence of a sync arc in an input or output port represents TRUE, and its absence represents FALSE. A sync arc between nodes $x$ and $y$ is shorthand for saying that $x$ and $y$ are in exact synchrony. Sync arcs are totally different from "connective arcs" in the graph, which represent synapses.

\subsection*{ AND gate }
\begin{figure}[h]
\centering
\includegraphics{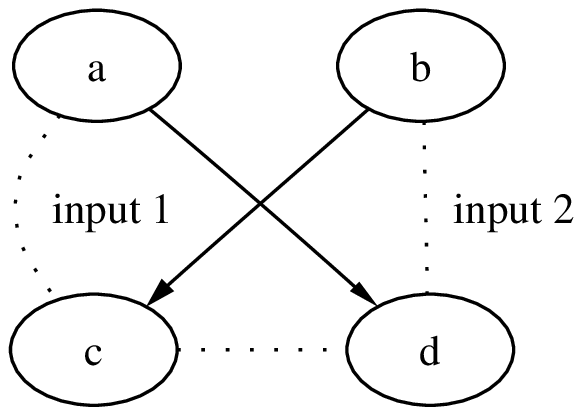}
\caption{Simple AND gate. The output of the gate is the potential sync arc at CD.}
\label{AND1}
\end{figure}

Fig. \ref{AND1} is a simple model of a logical AND.

This AND gate gets two input arcs. Neurons $a, b, c$, and $d$ all receive external input. Denote the sets of times at which they receive spikes  as $A, B, C$ and $D$. 

We will verify all four entries in the AND truth table.

\subsubsection*{ FALSE and FALSE is FALSE }
If input 1 and input 2 are both FALSE, this means that $a$ and $c$ are not synched with each other, and neither are $b$ and $d$. So we have 

$A \neq C 
B \neq D$

and because we can assume that any inputs which are not explicitly synchronized are not synchronized, we have

$A \neq B 
A \neq C 
A \neq D 
B \neq C 
B \neq D$

Now, by the connectivity diagram, we see that $C_{total} = C \cup B$, and $D_{total} = D \cup A$. Since set unions are distinct unless they have been forced to be identical, we may conclude that $C \cup B \neq D \cup A$. Hence, the total input to $c$ differs from the total input to $d$. So, the output of $c \neq$ the output of $d$. So, $c$ and $d$ are desynchronized; the output represents FALSE, as we wanted. 

We'll go through the other cases a little faster.

\subsubsection*{ FALSE and TRUE is FALSE }
If input 1 is FALSE and input 2 is TRUE, this means that $a$ and $c$ are not synched with each other, and $b$ and $d$ are synchronized with each other. So we have 

\begin{align*}
&A \neq C
\\B &= D
\end{align*}

By the connectivity diagram, we see that

\begin{align*}
C_{total} = C \cup B
\\D_{total} = D \cup A
\end{align*}

and we have

\begin{align*}
C_{total} = C \cup D
\\D_{total} = D \cup A
\end{align*}

Since set unions are distinct unless they have been forced to be identical, we conclude that $C \cup D \neq D \cup A$. Hence, the total input to $c$ differs from the total input to $d$. So, $c$ and $d$ are desynched, and the output represents FALSE, as we wanted. 

\subsubsection*{ TRUE and FALSE is FALSE }
Similar to the last case.

\subsubsection*{ TRUE and TRUE is TRUE }
If input 1 is TRUE and input 2 is TRUE, this means that $a$ and $c$ are synched , and $b$ and $d$ are synched. 

\begin{align*}
A &= C
\\B &= D
\end{align*}

By the connectivity diagram, we see that

\begin{align*}
C_{total} = C \cup B
\\D_{total} = D \cup A
\end{align*}

and we have

\begin{align*}
C_{total} = C \cup B
\\D_{total} = D \cup A
\end{align*}

$C_{total}$ = $A \cup B 
D_{total}$ = \begin{align*}B \cup A 
\\ \\  C_{total} = D_{total}
\end{align*}

Hence, the total inputs to $c$ and $d$ are equal, so $c$ and $d$'s outputs are equal, and they are synched. The output is TRUE, as desired.

\subsection*{ NOT gate }
\begin{figure}[h]
\centering
\includegraphics{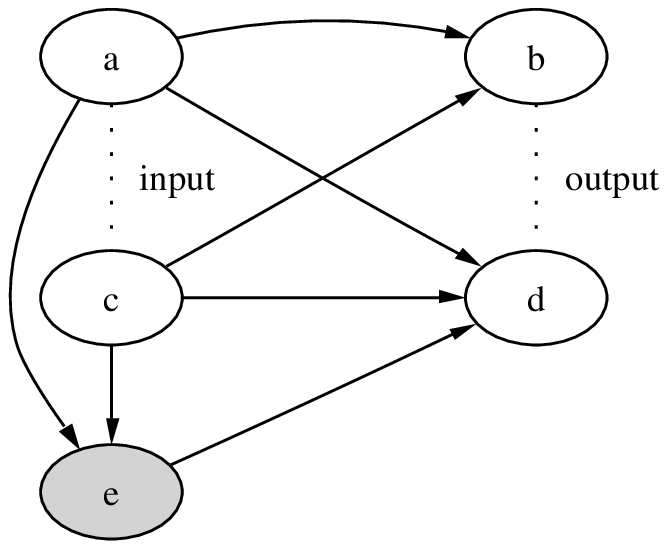}
\caption{Simple NOT gate. $e$ is a threshold unit with threshold 2. The output of the gate is the potential sync arc at BD.}
\label{NOT1}
\end{figure}

Fig. \ref{NOT1} is a simple NOT gate. If the input sync arc (AC) is present (i.e., if $a$ and $c$ are in exact synchrony), then the output sync arc (BD) will be absent, and vice-versa.

%TODO: c -> e???

If $a$ and $c$ are not sync'd, then $e$ does not fire (remember, $e$ is a threshold unit which will not fire unless it receives at least two synchronous inputs), and we have

\begin{align*}
B_{total} = A \cup C
\\D_{total} = C \cup A
\\B_{total} = D_{total}
\end{align*}
So, the total input to $b$ and $d$ are equal, so $b$ and $d$'s outputs will be identical and $b$ and $d$ are in sync. So there will be a sync arc in the output (a sync arc at BD). So if the input if FALSE, then the output is TRUE.

On the other hand, if $a$ and $c$ are sync'd, then $e$ does fire, and we have

$B_{total}$ = $A \cup C 
D_{total}$ = $C \cup A \cup$ E
$B_{total} \neq D_{total}$

So, the total input to $b$ and $d$ are not equal, so $b$ and $d$'s outputs will be different; $b$ and $d$ are not in sync. So there will be no sync arc in the output (at BD). So if the input if TRUE, then the output is FALSE.

Note that it takes this gate 2 time-steps, not just 1, to for a state change to get through it. This is because of the intermediate node $e$. This doesn't affect things because we could always add an extra delay hop to all of the other logic gates so that they would ALL take 2 time-steps.

\subsection*{ COPY gate }
\begin{figure}[h]
\centering
\includegraphics{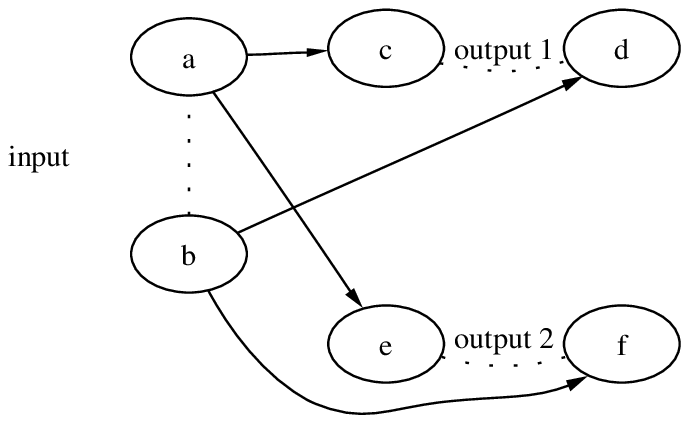}
\caption{Simple COPY gate.}
\label{COPY1}
\end{figure}

COPY is a 1-input, 2-output function; COPY(FALSE) = (FALSE, FALSE) and COPY(TRUE) = (TRUE,TRUE). Fig. \ref{COPY1} contains the obvious wiring\footnote{Note that the output node $c$ will also be synch'd with $e$, and $d$ with $f$. This doesn't matter, because remember that down the line if a signal gate receives these two signals on separate ports, that the input nodes on different ports that gate are assumed to have different "intrinsic properties" which desync those inputs.}.

\subsection*{ Turing-completeness }
Note that all we needed to show Turing completeness was NAND and COPY. We have an AND gate, and a NOT gate, so we can compose them to get a NAND gate. So, we have already shown Turing-completeness.

You may wish to see some alternate implementations of logic gates in Appendix \ref{moreSynchGates}.

\chapter{Some non-boolean logic gates implemented with synchrony} \label{nonboolSynchGates}

The boolean logic gates are all we need to show that the graph computation machine is capable of computing any computable function.

However, the exact synchronization state of a network carries additional structure that it might be interesting to use. Here are some additional gates which take advantage of that structure.

\subsection*{ The constant SILENT }
We have been dealing with a two-valued logic; synchronization between two nodes denotes TRUE, and no synchronization denotes FALSE. We could introduce a third truth value, SILENT, denoting a state when neither neuron is firing. We won't introduce a full basis of gates for dealing with silent, but rather will just give one to demonstrate the idea.

We introduce a new kind of neuron, called an \emph{inhibitory neuron}\footnote{We also need to strengthen our "no accidental synchronization" conditions. Now we require that each input port neuron emit at least one spike time which none of the other input port neurons do. This is so that we can't get accidental synchronization even if we subtract these different port signals from each other.}. An inhibitory neuron is essentially a "minus sign" with no internal dynamics. Unlike other neurons, the inhibitory's output is completely specified; its output is the same as its total input. Also, it's output connections possess a peculiar property. If $h_1\ldots h_n$ are inhibitory, and the $h_i$s' outputs connect to $x$, then $x$'s "total input" is redefined to be

\begin{align*}
X_{total} = X_{excitatory\ total} - H_1 - \ldots  - H_n
\end{align*}

For example, if $x$ gets input from $h$, $a$, and $b$, and $A = \{5,15\}$, $B = \{0, 10\}$, and $H = \{7, 10\}$, then $x$'s total input is calculated as

\begin{align*}
X_{total} = \{5,15\} \cup \{0, 10\} - \{7, 10\} 
\\= \{0,5,10,15\} - \{7, 10\}
\\= \{0,5,15\}
\end{align*}

\begin{figure}[h]
\centering
\includegraphics{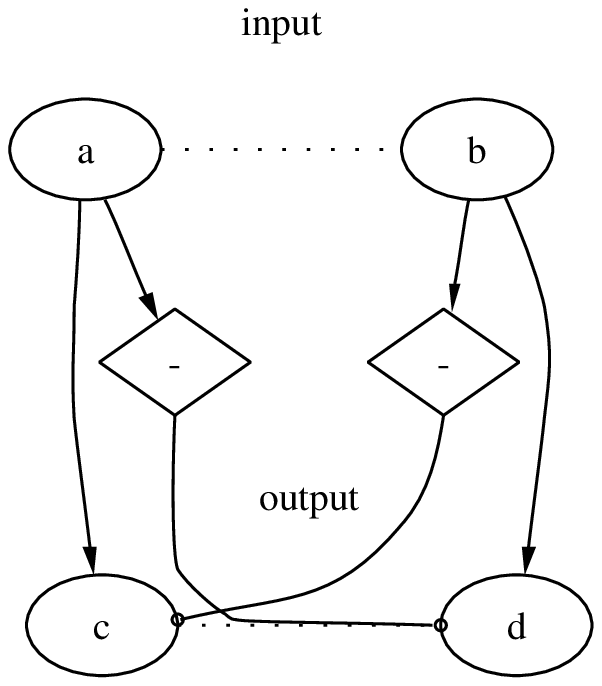}
\caption{A gate that takes FALSE to TRUE and TRUE to SILENT. The diamond-shaped neurons are inhibitory.}
\label{trueToSilent}
\end{figure}

Fig. \ref{trueToSilent} denotes a gate which takes FALSE to FALSE and TRUE to SILENT.  That is, as input, it gets two nodes. If those nodes are synchronized, then it emits nothing; otherwise, it emits an unsynchronized signal. 

\subsection*{ The gate REPLICATE-IF }
Fig. \ref{REPLICATE} is a gate which accepts 2 pairs of nodes, and emits 2 pairs of nodes with the following properties:

\begin{itemize}
\item[*] Each output pair corresponds to an input pair, and an output pair is synchronized iff the corresponding input pair is synchronized. 
\item[*] If both input pairs are synchronized, then all four nodes of the two output pairs are synchronized. Otherwise, there are not "extra" synchronizations\footnote{Assuming, as usual, that the input ports of the REPLICATE-IF gate do not have allow accidental synchrony between them.}
\end{itemize}

\begin{figure}[h]
\centering

\entrymodifiers={++[o][F-]}
\xymatrix{
a \ar[ddd] \ar[dddrrr] \ar[dddrrrr] 
& b \ar[ddd] \ar[dddrr] \ar[dddrrr] 
& *{} 
& c \ar[ddd] \ar[dddll] \ar[dddlll] 
& d \ar[ddd] \ar[dddlll] \ar[dddllll] 
\\ *{}
\\ *{}
\\e & f & *{} & g & h
}

\caption{The REPLICATE-IF gate.}
\label{REPLICATE}
\end{figure}

In other words, REPLICATE-IF simply copies the inputs to the outputs, UNLESS both inputs are TRUE, in which case it outputs two ports which are cross-synchronized in addition to being individually TRUE\footnote{This cross-synchrony is unusable with our other gates because we have specifically required the input ports of all other gates to abolish any cross-port synchrony. However, we could construct a whole other set of gates which is sensitive to cross-synchrony.}.

\chapter{More implementations of gates} \label{moreSynchGates}
\subsection*{ Another AND gate }
\begin{figure}[h]
\centering
\includegraphics{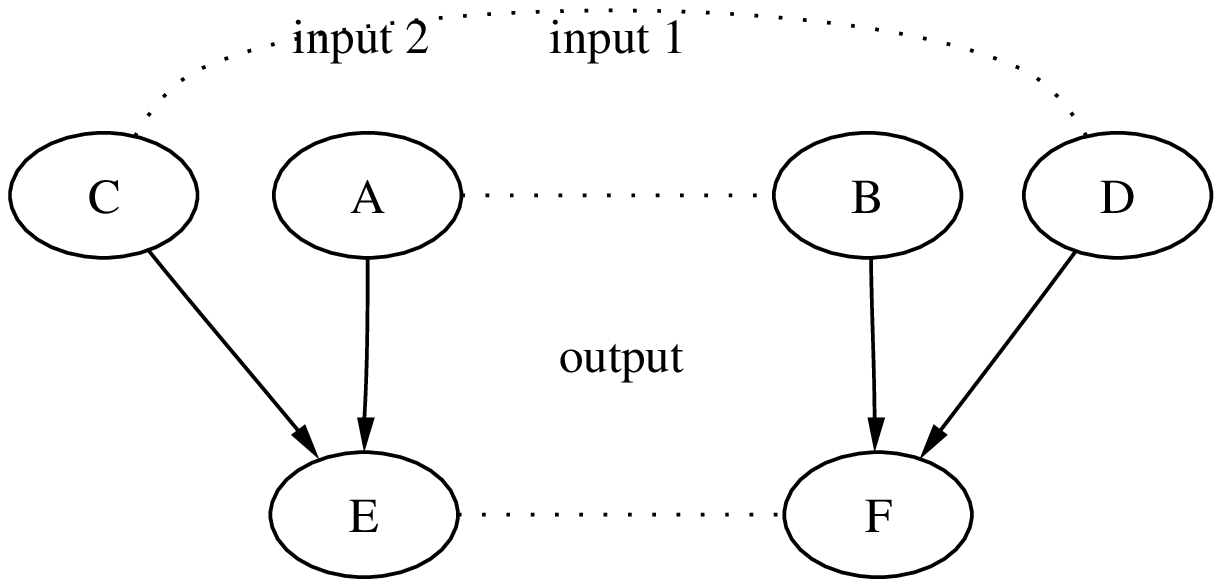}
\caption{The synch states (dotted lines) between AB and between CD are the inputs, and the synch state between EF represents the output. All connections are excitatory. The weight of the connections are .5. E and F do store activation.}
\label{AND2}
\end{figure}

Fig. \ref{AND2} shows an implementation of an AND gate. 

If A and B are sync'd and C and D are sync'd, then E and F receive the same activation on each timestep. Hence E and F will be sync'd; the gate takes (TRUE, TRUE) to TRUE.

If A and B are not sync'd, then E and F will receive spikes at different times. Similarly if C and D are not sync'd. Similarly if none of them are sync'd\footnote{Because by our "no accidental synchronization" conditions, we know that $A \cup C$ is never equal to $B \cup D$ unless it has to be.}. So the gate takes everything else to FALSE.

\section*{ A different model of neurons }
We have a different toy physics model for the next gate. 

Let each neuron have an "activation state", and a threshold parameter. The activation state may be stored, i.e. it may build up over time, or it may be unstored, i.e. it dissipates so quickly that it does not affect the next timestep. Any input is simply added to the activation state, and the neuron fires when it goes above threshold. For simplicity, we can set the threshold to 1. Some neurons also slowly accumulate activation on their own at the rate of $1/T$ units per timestep. Outgoing spikes acquire the strength of the synapse over which they travel. 

Some neurons also have a "relative refractory period" implemented as follows: even if the activation reaches 1, a neuron will not fire if it has fired within the last $H$ timesteps. However, if the activation reaches some constant $A$, then it fires regardless.

\subsection*{ Another NOT gate }
\begin{figure}[h]
\centering
\includegraphics{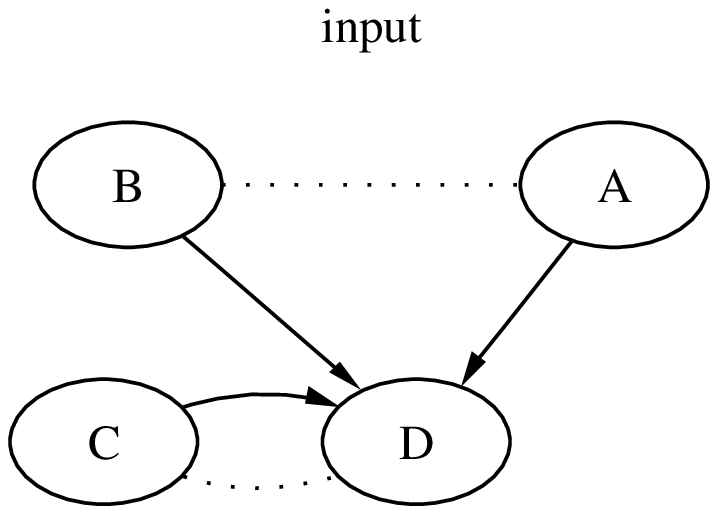}
\caption{The synch state (dotted line) between AB is the input, and the synch state between CD represents the output. All connections are excitatory. The weight of the A$\to$D and B$\to$D connections are .8. The weight of the C$\to$D connection is 1. D has a relative refractory period of 8 timesteps. The activation needed to overcome the refractory period is 1.5. C intrinsically oscillates; it fires every 10 time steps. D has no intrinsic activity. D does not store activation.}
\label{NOT2}
\end{figure}

Figure \ref{NOT2} shows an implementation of a NOT gate. 

Consider the case when units A and B are not sync'd. In this case, the firing of A and B will rarely be simultaineous, and so A and B rarely cause D to fire (because the weights of the A$\to$D and B$\to$D connections are only .8, and because 8/10ths of the time D is in its relative refractory period from the last time C caused it to fire). However, D will fire whenever C fires, because the strength of the connection from C to D is 1. So B and D will fire in synchrony (almost) every 10 time units. So this gate takes FALSE to TRUE.

Now consider when units A and B are sync'd (with a period close to 10). In this case, A and B will fire simultaineously, and each time they do, D's threshold will be exceeded, even if it is in the relative refractory period, and D will fire. Sometimes C will also cause D to fire, but this will be rare, because most of the time D will be in its relative refractory period from the last time that A and B caused it to fire, and the strength of the C$\to$D synapse in not sufficient to overcome that.

\section*{ A third model of neural synchronization }
Here's yet another model of neural synchronization. This one isn't as well-developed. 

In this model, synapses have different strengths, and that neurons will become "enslaved"\footnote{Many would say "entrained", but I think that word is confusing} by synaptic input with sufficiently large strength, at which point they fire in the same pattern as that input. If there are multiple kinds of inputs, either they cause dissonance and the neuron is not enslaved by anyone, or, if the strength of one set of inputs is overwhelming, the neuron becomes enslaved by that pattern.

Reciprocal feedback is allowed, in which case the participating neurons tend to enslave/match each other in the absence of other input.

The neurons also fire in the absence of input; they each have a "natural cycle time" $T$ such that they will fire whenever they have not fired for the last $T$ timesteps.

We assume that it takes no time for a signal to propagate between these neurons\footnote{More accurately, the propagation time is small enough such that two neurons that fire one after another can be considered to be "in synchrony" by downstream neurons that integrate over at least very short periods of time}. 

However, I haven't created a precise implementation of the above qualitative model. The gates I'll describe are explained in terms of this model, but since the model itself in informal I haven't formally verified that all of them actually work. I think that something like the previous more numerical model could support this qualitative scheme, if neurons activation was stored and built-up over time, but I haven't tested it.

\section*{ Yet another NOT gate }
\begin{figure}[h]
\centering
\includegraphics{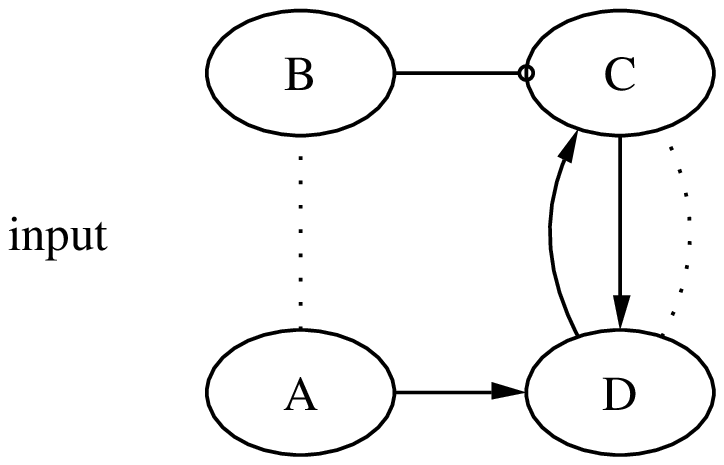}
\caption{The synch state (dotted line) between AB is the input, and the synch state between CD represents the output. Arrows represent excitatory connections. The B to C connection is inhibitory. }
\label{NOT3}
\end{figure}

This gate, Fig. \ref{NOT3},  requires inhibitory synapses to actually SUBTRACT "activation" from their targets when they fire.

This circuit is designed so that C and D's reciprocal feedback strongly inclines them to synchrony. Only by acting together can A and B disrupt that synchrony. 

When the input is FALSE, AB are out of sync, and the B$\to$C and A$\to$D connections are too weak to overcome the recriprocal synchonry of C and D. So C and D are synchronous, and the output is TRUE.

When the input is TRUE, AB are in sync. Each time A fires, it adds activation to D, pushing the next time that D will fire sooner. Simultaineously, B fires, but since the B$\to$C connection is inhibitory, it SUBTRACTS activation from C, which pushes the next time that C will fire LATER.

So, each time A and B fire, they work to push C and D out of phase as much as possible. We set up the connection strengths so that when this happens, the disruption is strong enough to destroy C and D's natural synchrony.

So, when the input is FALSE, the output is TRUE.

\subsection*{ A NAND gate }
\begin{figure}[h]
\centering
\input{easyLatexGraph16268.tex}
\includegraphics{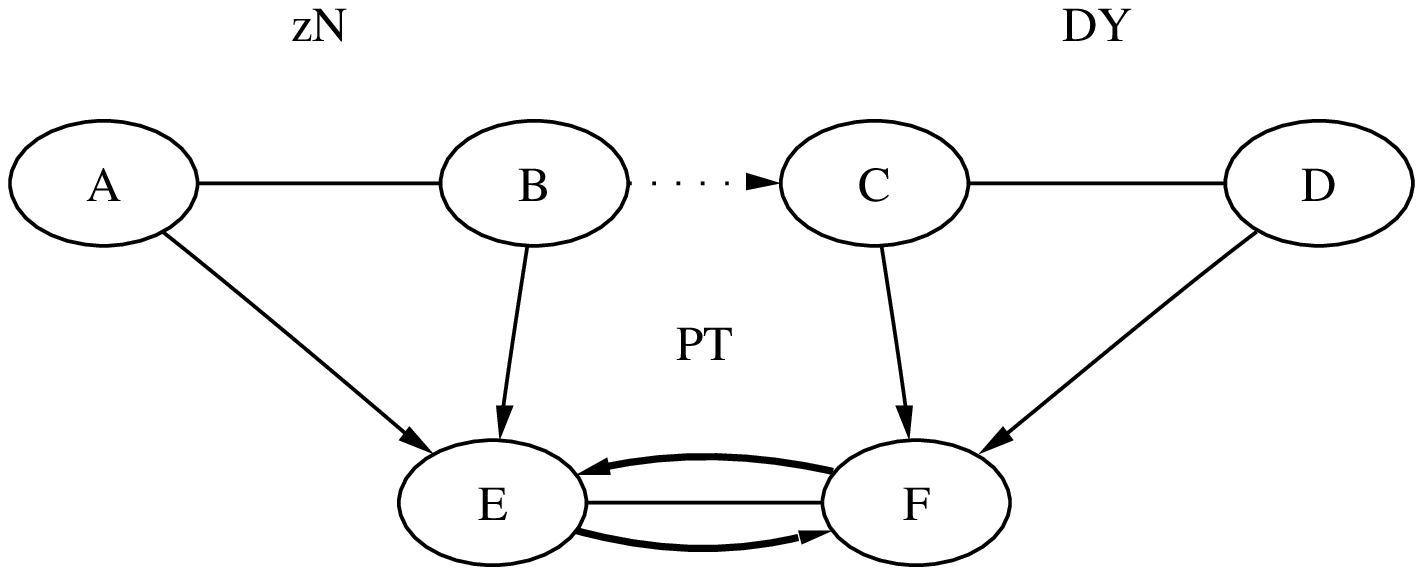}
\caption{A NAND gate. The first input is the synch state of AB, the second input is the synch state of CD, and the output is the synch state of EF. There are strong reciprocal connections between E and F. There are weaker connections from A and B to E, and from C and D to F. There is a weak inhibitory connection from B to C.}
\label{NAND}
\end{figure}

Fig. \ref{NAND} is a NAND gate.

This is a NAND gate, which means that E and F are always synchronized except when AB is synchronized and CD is also synchronized. 

When the input is (FALSE, FALSE), that is, when neither AB nor CD are synched, E and F's input is dominated by their own reciprocal connections, and they synchronize with each other. So the output is TRUE.

When the input is (TRUE, FALSE), that is, when AB is synched and CD is not, then the synchronous input of A and B into E is sufficient to "enslave" E, and E becomes synchronized to A and B. Through it's connection to F, however, E also brings F into synchrony with itself. So A,B,E, and F are all synchronized with each other. So the output is TRUE.

When the input is (FALSE, TRUE), the situation is similar.

When the input is (TRUE, TRUE), that is, when AB is sync'd and CD is sync'd, then the weak inhibitory connection from B to C ensures that they are not in synch with each other (that is, it prevents all four of them from firing in the same pattern; so we have A and B firing with one pattern, and C and D firing in a different pattern). A and B's synchronous input to E enslaves E, and C and D's synchronous input to F enslaves F. So E and F are no longer in synch. So the output is FALSE.

\chapter{A NAND gate using cycle computation}\label{appCycleNAND}

\begin{figure}[h]
\centering
\includegraphics{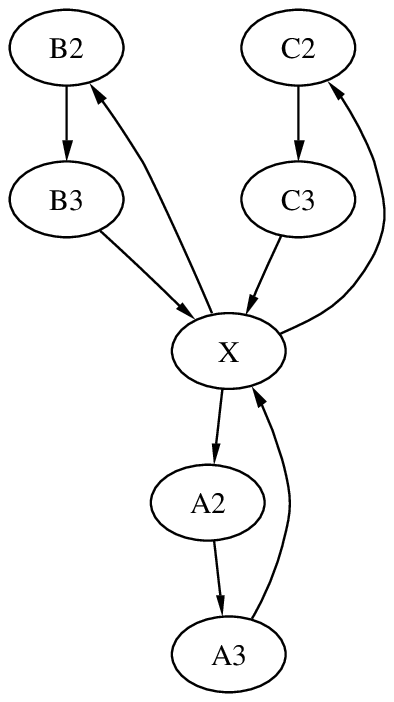}
\caption{A NAND gate using cycle computation, in a model where each cycle and each node has a set of allowed oscillation/resonance frequencies}
\label{cycleNAND}
\end{figure}

I haven't fully specified a toy physics model for cycle computation but here's the beginnings of one. Note that a signal in a cycle will go around the loop. Perhaps you could say that each node, and each cycle, can only support certain frequencies of cycling. If there are also intrinsic oscillators, then here is a NAND gate (see Fig. \ref{cycleNAND}.

Let node X be a member of cycles A, B, and C. A and B are inputs, and C is the output. Cycle A can only oscillate at frequency $w_1$, and cycle B can only oscillate at frequency $w_2$. Cycle C can oscillate at either $w_1$ or $w_2$. Node X can oscillate at either $w_1$ or $w_2$ or both at once (i.e $w_1 w_2$)). 

There is also some node in C which is an intrinsic oscillator, i.e., it is always trying to "jump-start" a signal in cycle C if there is no signal going around.

Now, if both cycle A and cycle B are active, they won't conflict with each other, but no signal can circulate in cycle C, because node X can't do $w_1^2 w_2$ or $w_1 w_2^2$. So, cycle C must be inactive. So, the gate maps (TRUE, TRUE) to false.

For any other set of inputs, though, there is at least one available frequency for C to cycle at. And, since it has an intrinsic oscillator, it will achieve that frequency. So, the gate maps everything else to TRUE. So, it is a NAND gate.

\chapter{Turing machines, and criteria for Turing-completeness of graphical computation} \label{appTuring}

\subsection*{ Background }
Basically, a Turing machine is something that, if you spent enough time figuring out how to program an emulator in it, could run computer programs. Given enough computing time and the proper programming, a Turing machine can emulate all known computing architectures\footnote{In fact, a Turing machine can simulate discretized Newtonian physics, and maybe quantum physics, too, although there's disagreement there. Some question whether a Turing machine can emulate a conscious entity, or emulate the brain, though.}\footnote{However, a Turing machine cannot efficiently emulate all computing architectures; there are some programs for which the time complexity of running the program on a Turing machine may be significantly greater than the time complexity of running that program in a specialized architecture designed for that program. Specifically, if the length of the input is known when the machine is designed, then sometimes we can design it to run significantly faster. This sort of optimization is referred to as a "non-uniform model of computation", as opposed to "uniform Turing machines", which must work for any input size (see Wegner, "The Complexity of Boolean Functions", http://eccc.uni-trier.de/eccc-local/ECCC-Books/wegener\_book\_readme.html). Hopefully this sort of thing will turn out to be a minor detail that we can ignore for most purposes.}. Most computing architectures can be proved to be capable of emulating a Turing machine. Therefore, there is a large class of devices which, given enough time and programming, can theoretically emulate any other device in the class. When this happens, the devices are said to be "computationally equivalent". Most computing devices are computationally equivalent to Turing machines.

Clearly, a human is at least as powerful as a Turing machine, because, disregarding time and memory limitations, it is possible for us to verbally emulate the execution of a computer program in our heads. We can't imagine a physical implementation\footnote{Heck, I can't even imagine a theoretical construct with more power} of something which has more computational power than a Turing machine. Therefore, when looking for architectures that the brain might use to compute, one criterion that we might use is Turing-equivalence. 

Basically, a Turing machine is something that, if you spent enough time figuring out how to program an emulator in it, could run computer programs. Therefore, it can implement boolean logic, it can do arithmetic calculations, it can execute operations conditionally, it can loop, and, crucially, it can get execute an "infinite loop". Computer programming languages which can do basic arithmetic, and which have an IF statement and a GOTO statement tend to be Turing-complete. 

\subsection*{ Checking for equivalence }
I did not rigorously prove that any of the architectures were Turing-equivalent. This might be a good thing to do sometime. For now, though, let's look at some less rigorous critera for a computational architecture to be as powerful as a Turing machine.

Note that although the systems we will study have usually have finite memory\footnote{but not necessarily; we could have a graph computation machine which accepts graphs of any size, and which is allowed to add arbitrary nodes and edges during the computation}, technically they would have to have infinite memory in order to be a Turing machine. We will overlook this fact\footnote{You have to watch out when you overlook this. If you know that the input to some function will never exceed some finite size, you could implement the function as a lookup table (perhaps the lookup table could output a special value to mean "infinite loop" if you want to require the resulting machine to have that behavior). Lookup tables can't really branch or do GOTOs, but they can still get the correct result. We want to disallow this sort of thing. A machine that can be programmed to compute any finite function, but only if you first precompute the lookup table and use that as the "program", isn't very useful. 

With Turing machines, you avoid this case by requiring the machine to have the capability to run finite-length programs which can handle arbitrarily large inputs. Intuitively, the situation is simple. We'll just avoid machines which can't be extended to handle larger memories/input sizes in a straightforward way. 

Formally, if you attempted to say "a lookup table can emulate a Turing machine, provided you set it up right", you would find that it was very difficult to figure out how to set up the lookup table. First, you would have to know, for each input that the table might get, whether or not a Turing machine would halt on that input; an uncomputable task. Second, even disregarding that, it would take a ridiculous amount of time and space to find the right "setup" for the lookup table because you'd have to execute a Turing machine for every possible input. 

Formally, then, we can require that the complexity of figuring out how to extend our system to be ready for different maximal input length must be feasible. Technically, we say that we want only "uniform" computational models, rather than "non-uniform" ones (i.e. for different input sizes, their design stays mostly the same). A formalization of uniformity for the special case of systems of boolean circuits can be found in Chapter 9, section 8 of  Wegner, "The Complexity of Boolean Functions", http://eccc.uni-trier.de/eccc-local/ECCC-Books/wegener\_book\_readme.html

}.

\subsubsection*{ Boolean Logic }
The first thing we should look for in a system is a capability for boolean logic. Let's say that we can figure how to do some boolean operations with a system, and we wonder if, by combining these operations, we can generate the rest. Post's Theorem gives a set of necessary and sufficient conditions\footnote{ The theorem is: A necessary and sufficient condition that a set of operations form a basis for the nonzeroary Boolean operations is that it contain operations that are respectively nonaffine, nonmonotone, nonstrict, noncostrict, and nonselfdual. }\footnote{Vaughan Pratt's CS353 notes, http://boole.stanford.edu/cs353/handouts/book3.pdf} for a basis of operations to be capable of generating all boolean operations. A NAND gate is such a basis. In addition, it is necessary to be able to compose operations and to copy bits. So, composition, copying, and the capability to compute a NAND gate are necessary and sufficient conditions for a system to be capable of computing boolean functions\footnote{Actually, Post's theorem is typically set up in such a way that composition and copying is assumed. However, clearly the set of functions which a system can compute must be closed under composition of boolean functions, if the system is to be capable of all boolean functions. The 1 input, 2 output boolean function COPY is also a boolean function and hence must be expressible. So, these properties are necessary. Post's theorem shows sufficiency.}

\subsubsection*{ Memory }
After we figure out how to compute Boolean functions, we would like our system to be able to store memory. A Turing machine has memory in the form of its memory tape; an linear sequence of memory elements which hold their state over time and whose state can be altered by the machine's head.

Memory storage is not quite as important for our purposes, because even if we find a system which cannot store memory, we might assume that the brain provides memory storage elements as a "primitive" external to the part of the system that we are modeling. So, we should feel free to add memory storage primitives to any of our models.

However, this won't be a problem for most of our models. Anytime we have a system of boolean gates wired together in a circuit, where it takes time for a signal to propagate through the circuit, we can build a memory element using recurrent connections. See Appendix \ref{appMemory} for how.

\subsubsection*{ Emulating movement of the Turing machine head  }
A Turing machine has a "head" which moves about on a memory "tape", reading and writing. But what if our system doesn't have moving parts?

The "head" could be constructed by having a bunch of adjacent, identical "blocks" of circuitry. Each block contains a "head-template", that is, circuitry which can compute the actions of the Turing machine "head" when it is at that location. 

\includegraphics{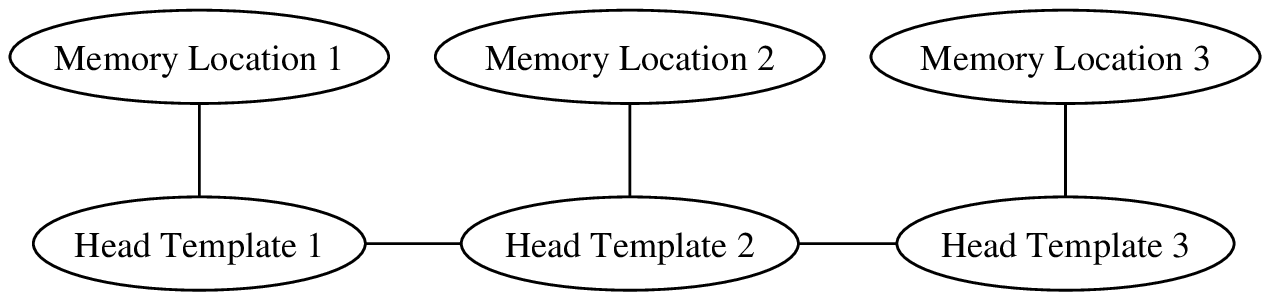}

Say that the head is in position 1. This is emulated by having the head template at position 1 being "active", and all of the other head-templates "inactive". 

Say that the head wants to move right to position 2. Instead of actually moving anything, the head-template at position 1 could signal to the head-template at position 2 to activate. It would also pass the internal state of the head to position 2. The head-template at position 1 then deactivates.

So, you don't need to have moving parts to build a Turing-machine head if you can instead have a large number of repeated circuit blocks adjacent to each other. 

Although it is unlikely that the brain directly implements something like a Turing machine (which has a read/write head, possessing internal state, that moves among an array of memory locations), it should be noted that biology is particularly good at building large arrays of identical (or similar) parts. For example, neurons, cortical columns, etc. So, it might be a good idea to remember situations, such as this one, where such a geometry turned out to be useful.

Another way that one might imagine emulating head motion is to have a single CPU that accesses memory locations in some sort of RAM. The "position"/current memory location of the Turing head could be encoded as one of the registers in the CPU. There are a couple ways to realize this, but one of the simplest involves a tree-like structure whose root is the CPU and whose leaves are the memory locations; the structure contains intermediate nodes which process and route the CPU's request to read/write from a given address, making sure that the data goes to and from the correct memory location. See Appendix \ref{appMemory} for an example. This structure seems to be even further removed from the brain, however.

\subsubsection*{ State transition table in the Turing machine head }
The Turing machine head has a notion of internal state, and a small, memoryless state-transition table that changes the state depending on the current state, and on the value of the current memory location.

Internally, the head just has a small, memoryless, state-transition table, which we know that we can implement because we already know that our system can compute boolean logic.

\subsubsection*{ Loops }
At this point we have emulated a Turing machine (although we have not shown this formally), so we should be able to run loops, gotos, etc.

\subsubsection*{ Summary of Checking for Equivalence section }
So, if we can implement some logic gates including a NAND gate and a COPY gate, and if we can hook up the gates to each other in any configuration, and if it takes time for signals to travel from one gate to the next, then we can emulate a Turing machine.

Note, however, that had been looking at BASIC-like "programs" instead of "circuits", that it would have been possible to create a programming language that could compute boolean logical expressions, and which had memory, yet did not have any loops or GOTOs and which was not Turing-complete. For example, a language might have "IF", "AND", "OR", and "NOT", and memory assignment (even with fancy data structures), but nothing else. In that case, we could still write programs like:

\begin{verbatim}
10 new clause1[2];
40 clause1[0] = INPUT1;
60 clause1Truth = 0
100 v1 = 0;
110 v2 = 0;
120 if ((clause1[0] = 0) AND (v1=1)) THEN clause1Truth = 1
130 if ((clause1[0] = 1) AND (v2=1)) THEN clause1Truth = 1
120 if ((clause1[1] = 0) AND (v1=1)) THEN clause1Truth = 1
130 if ((clause1[1] = 1) AND (v2=1)) THEN clause1Truth = 1
140 IF (clause1Truth = 1) THEN 
150    print "When v1=0, v2=0, your clause is true."
160    halt
200 v1 = 0;
210 v2 = 1;
220 if ((clause1[0] = 0) AND (v1=1)) THEN clause1Truth = 1
230 ...
\end{verbatim}

but we cannot enter an infinite loop (or any kind of loop). How is it that our access to boolean operations does not give us the power to emulate a Turing machine, the way that it did in the circuit case? The problem is that, before, we could wire together the gates themselves into recurrent loops. In this simple programming language, although we can form \emph{linear expressions} using boolean operators, we cannot form loops with them. In addition, one might note that the flow of time has different effects. Our logic gate circuits were each active on every timestep; a logic gate could be "in the thick of things" at the beginning of processing, and then still have a role later on. However, in this simple programming language, once a line of code has been passed, it has no effect on the program afterwards; and, a line of code is executed each time step until the program halts, no matter what. 

This sort of programming language is like a logic gate circuit which is constrained to be acyclic.

\newpage
\chapter{Building a memory from boolean gates which take time to compute} \label{appMemory}
Assume that it takes one unit of time for a signal to pass through each gate. Imagine a circle of gates strung together, where each gate is an OR gate, whose inputs are the previous gate, and the constant value 0:

\includegraphics{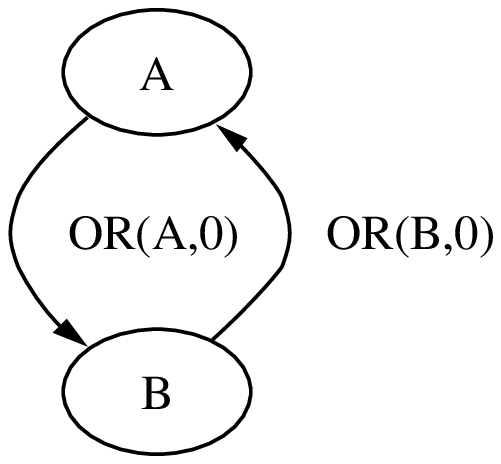}

If the value of both of $A,B$ are 0, then the system is in a stable state and nothing will change on the next time step. This represents the storage of bit "0" in memory. On the other hand, if all of the nodes have value 1, then they will remain 1; this represents the storage of bit "1".

Now, imagine that there are 1-bit "I/O ports", interfaces to the memory element. There is one port called "read", one called "erase", and one called "write". The value of the "read" port is always equal to the contents of the memory. If the "erase" bit is set, then a "0" is written into memory. If the "write" bit is set, a "1" is written into memory. This can be implemented as follows:

\includegraphics{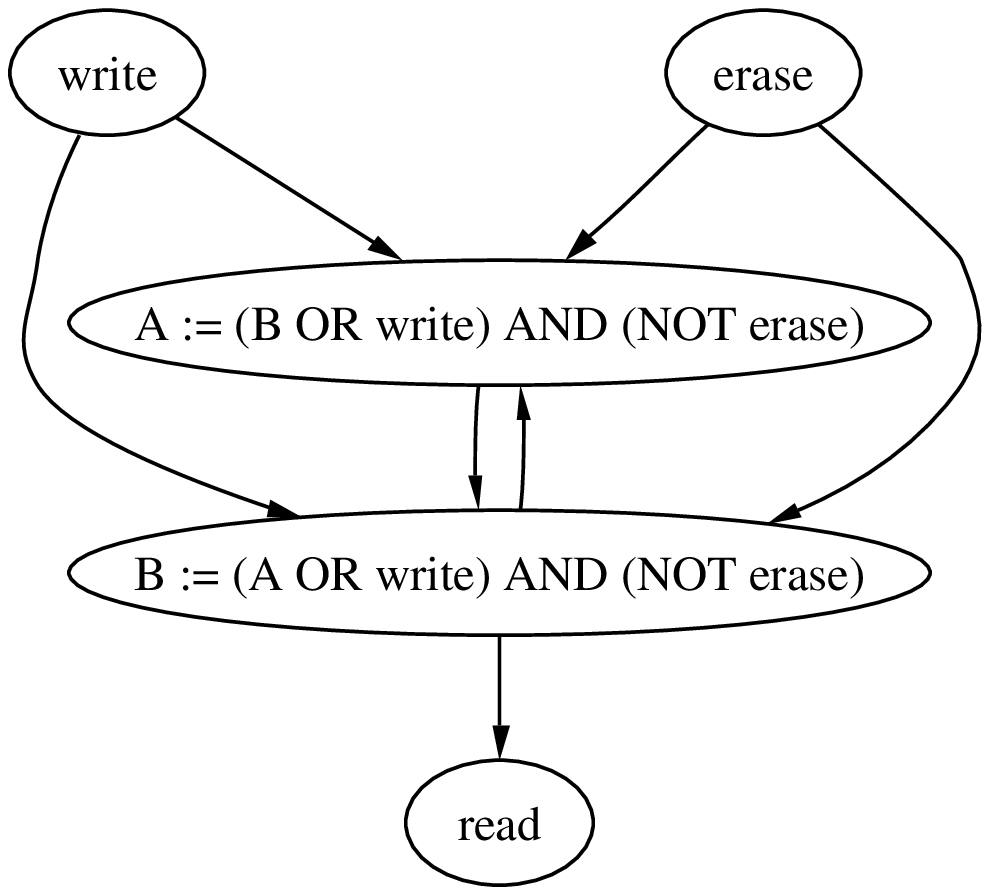}

% "read" label="read := A"; dropped for a nicer looking graph!

% (the lines connecting "read", "write", and "erase" to $A$ and $B$ are not shown; but the effects of "write" and "erase" on $A$ and $B$ can be seen in the labels for those nodes) % (they're shown now!)

In order to read a bit from the memory, you just look at the state of the "read" port. In order to write a bit to memory, you first set the "erase" port to 1, then you set the erase port back to 0, then you set the "write" port to the value that you want to write. After this, the memory will hold the value that you put in it until the next time you activate "erase" or "write". 

In order to make the memory element easier to use, we might put a "frontend" on it to handle the "erase, then write" sequence:

\includegraphics{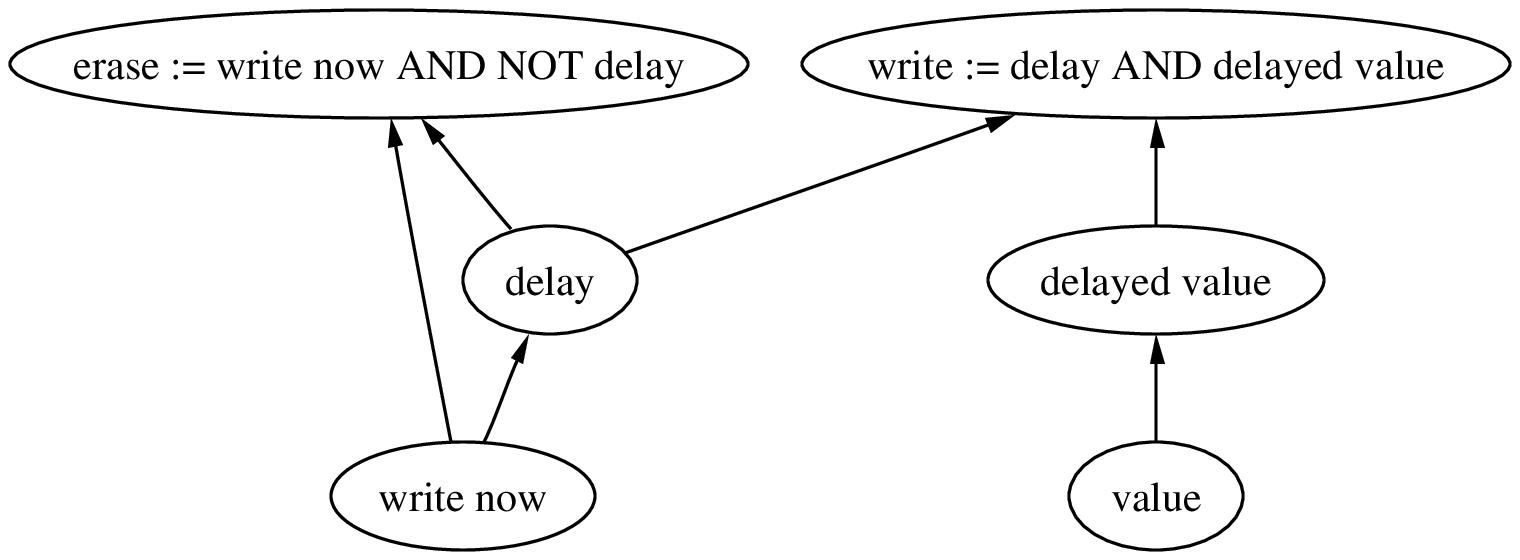}

Now, to write to memory, you load your value into the "value" port, and then set "write now" to 1, then set "write now" back to 0 later on. As long as you do not set "write now", the value of "value" may change arbitrarily without affecting memory. 

If "value" holds a value that you wish to commit to memory, and you activate "write now" at time $t$, then "erase" will be activated at time "t+1", and then at time "t+2", "erase" will be deactivated and "write" will hold the value that you are writing into memory. 

\subsubsection*{ Clocks }
Note that a clock circuit also be built in a similar manner, by having a signal circulate around a loop:

\includegraphics{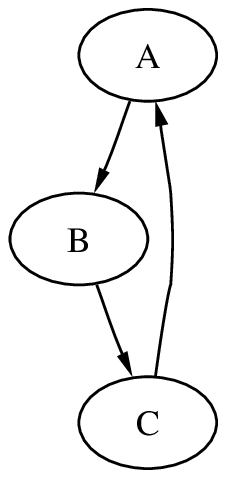}

\subsubsection*{ RAM }
Let's say that we have a finite amount of memory, and a single CPU, and we want to give that CPU read/write access to all of the memory (note that the structure could be duplicated to give multiple CPUs simultaineous access to the same memory locations). Note that this is probably not very close to how the brain operates. Still, here's one way to do this.

Say there are 8 memory locations. An address is specified with 3 bits. Here is how the memory would go to and from a given CPU region.

The CPU has access to a memory bank interface node. There is a 7 bit interface. 3 bits for addressing, one "set memory to this value" bit, one active memory bit for reading and one for writing, and one bit that gets the results of memory reads.

\includegraphics{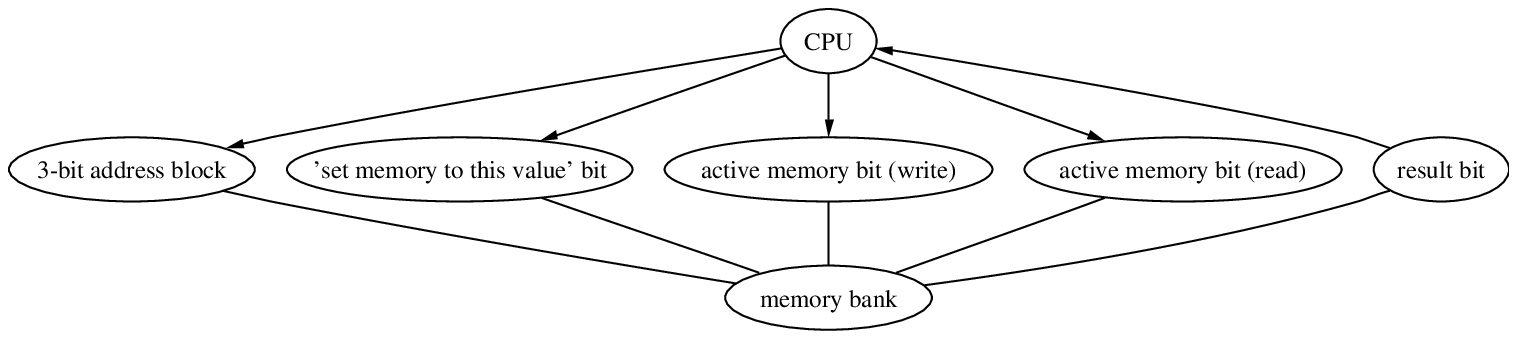}

This interface node sits on top of two memory banks, each of which contain 4 bits of memory. The left memory bank has addresses 000-011. The right memory bank has addresses 100-111. Each of them presents the 8-bit memory bank interface node with their own interface node.

\includegraphics{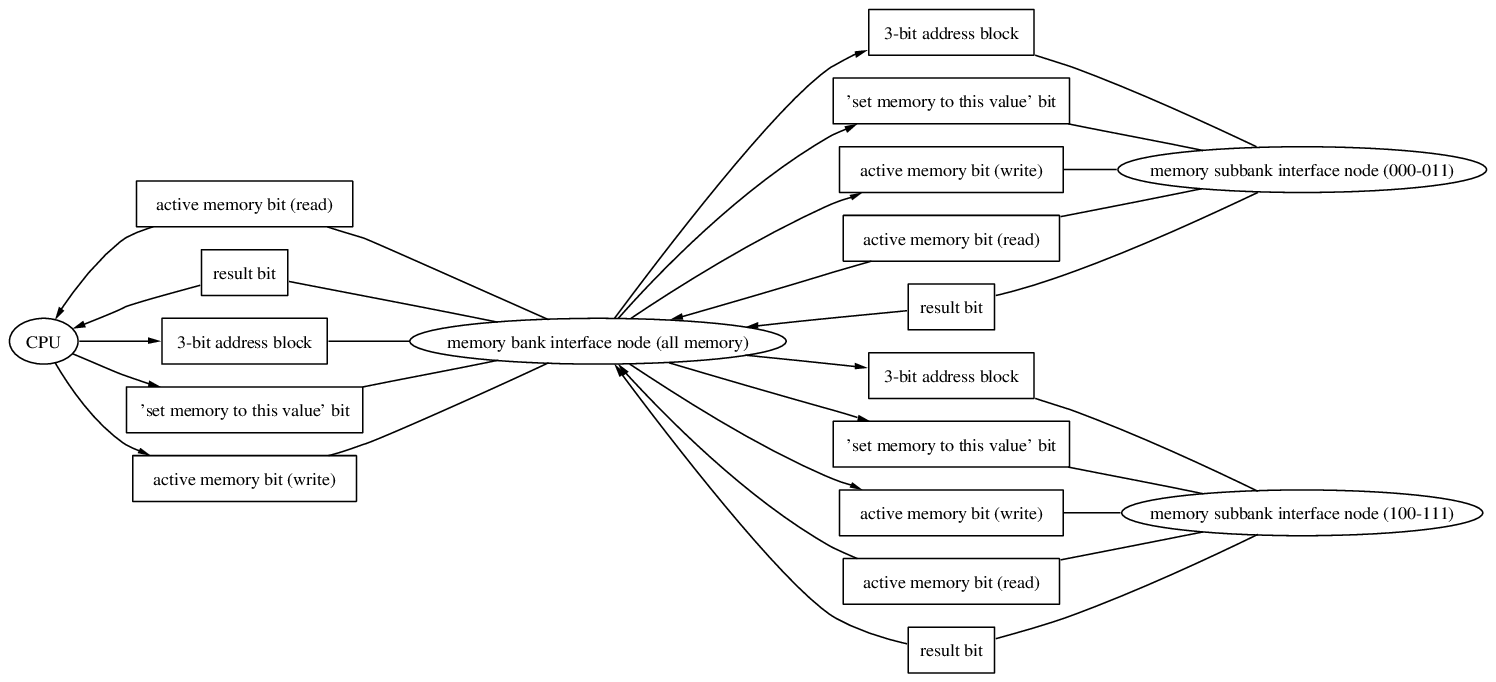}

This hierarchial system repeats until we get down to memory banks containing only a single memory location (the actual memory).

\includegraphics{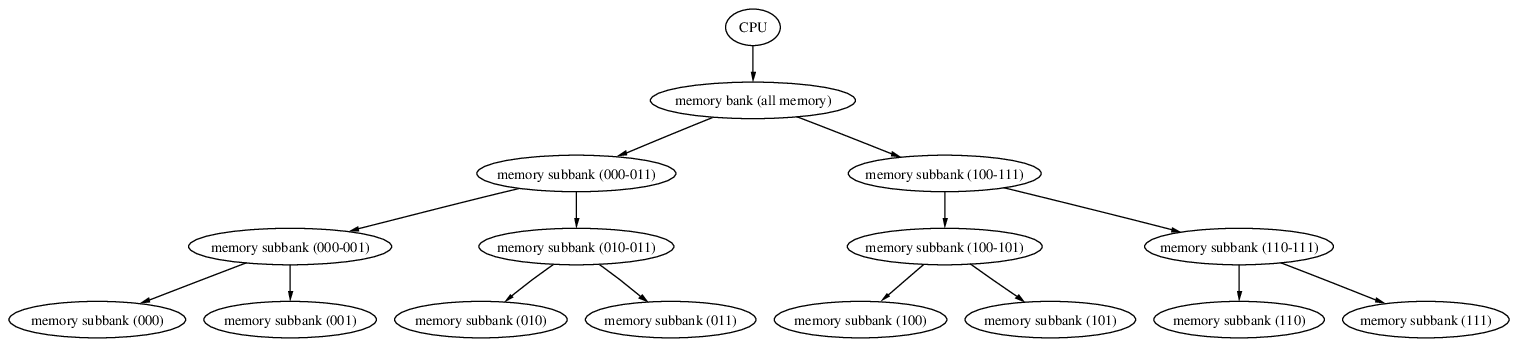}

Say you want to write a '1' to memory location 5. You set the address block to "101", and set the "set memory to this bit" to the thing that you want to write. Now you turn on the active memory bit. Through a system of copies, the memory subaddress  "01" is propagated to two units sitting above the memory interface

\includegraphics{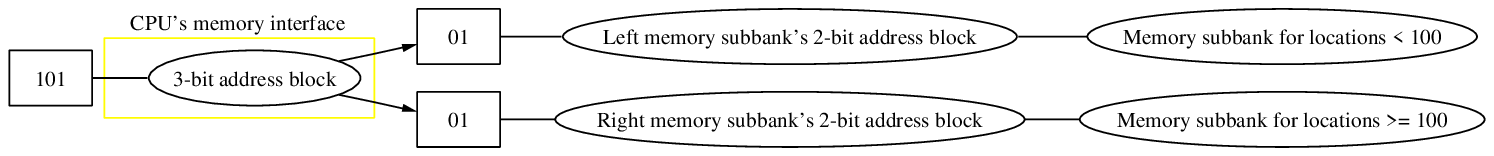}

So, now we have two "memory banks" who each have the subaddress "01" stored in leaf nodes. They need to decide which of them will process the request. Each bank has a constant written into it, telling it which bank it is (i.e. the "left" bank, "0" or the "right" bank; the value will be 0 or 1 respectively). They compare this value to the first bit of the address in the memory interface node underneath. So, memory bank "$<$ 100" gets a "0" result for the compare, and memory bank "$\geq$ 100" gets a "1" result.

Call this result the "am-i-the-right-memory-bank" result. This bit is then ANDed with the "active memory" input bit. The result (which is "1" for bank "$\geq$ 100" in the example) is stored in the "active memory bank" bit for the next two memory banks down the line.

There is a circuit which turns off the original input active memory bit at the memory interface node whenever any of the "active memory bank" bits go on.

So, here is what is happening so far:

\includegraphics{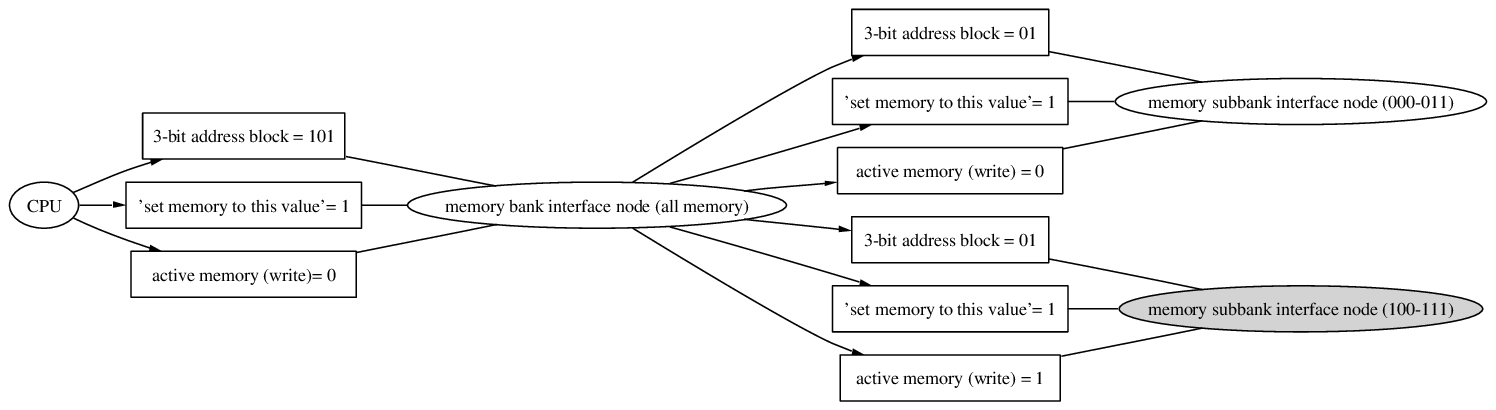}

Both the left and right memory banks have received copies of the bit to be written to memory, and the right hand portion of the address. But, because the first bit of the address was "1", only the right hand memory bank is active (as expressed in its active memory bit).

Now, on top of these two memory nodes are four more nodes (a binary tree). They are identical to the first two nodes. The values are passed on and modified the same way as before.

So, recursing, we see that eventually we get a memory node for the location "101", which has the "active memory bank bit" set (and which is the ONLY memory location with that bit set), and which has the value that we conveyed.

Reading the memory location can be done by having a second network that works somewhat in reverse. When the "active memory (read)" signal is set, the "active memory (read)" bits in the reading nodes become set to correspond with the correct memory bank. This all works similarly to the writing system. When the memory storage level is reached, there is an assignment from the memory location to an output pipeline. The output pipeline to hooked up the next node below this memory location as follows: it is ANDed with the "active memory (read)" bit, and then the result is assigned to the node below.

In this way, the result is conveyed to the requesting node (the root of the tree, at the bottom). However, if the result is 0, the requesting node can't distinguish between a 0 read, and between the result having not arrived yet. So, there is another pipeline from the top node to the bottom, which sends a "1" down when a top node becomes active.

The idea is that in between reads, the "go ahead and read" signal is turned off, which propagates up and turns off the upper memory nodes, and then their "result arrived" flags go off. So, when the next read is made, the "result arrived" flag really doesn't go on until the memory address reaches the top (and, an extra delay should probably be inserted in case there are delays in the signal getting down).

\chapter{Miscellaneous} \label{appMisc}
\section*{ More complicated ways to model synchronization }
There are plenty of other conceivable ways of modelling synchronization. We could even do physics; label the nodes with frequencies and other internal state values, and include a physical simulation in the "update rule". Too much of this negates the value of the project, though, as the goal is simplify the physics into something more manageable, while retaining just enough detail to model of the computation.  Here are some intermediate things that we could do:

\begin{itemize}
\item[*] Assume that all of the neurons are firing at the same frequency, and then label the arcs with the phase differences. 
\item[*] Ignore the phase and the actual frequency values, and let the arc labels represent the ratios between frequencies.
\end{itemize}

\part{Another EasyLatex ad}
\section*{ Just another plug for easylatex! With EasyLatex you can produce this$\ldots$  }
The identity matrix is the matrix for which $a_{ij} = 1$ when $i = j$, and 0 everywhere else. For example, in three dimensions, \begin{align*}I = 
 \left[ \begin{array}{lll}
   1 & 0 & 0
\\ 0 & 1 & 0
\\ 0 & 0 & 1
\end{array} \right]
\end{align*}
.

Here is a column vector: 
\begin{align*}
\left[ \begin{array}{l}
   0
\\ 1
\\ 0
\end{array} \right]
\end{align*}
.

\begin{align*}x \mapsto y_1\ldots y_n 
\\ \\ \frac{1}{3} 
\\ \\  1 + 1 &= 2
\\1 &\leq 3 - 1
\\&\leq 5
\end{align*}

\psfrag{TDH}[cc][cc]{$v_1 v_1$}
\psfrag{M_D}[cc][cc]{$v_2 v_2$}
\psfrag{mxE}[cc][cc]{$v_3 v_3$}
\psfrag{c2}[cc][cc]{$v_1$}

\includegraphics{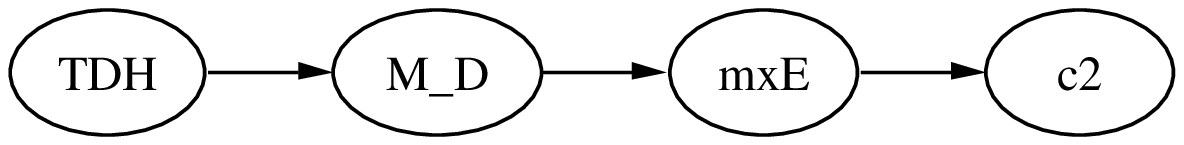}

\section*{ just by typing in this$\ldots$  }
\newpage

\begin{verbatim}
The identity matrix is the matrix for which a_{ij} = 1 when i = j,
and 0 everywhere else. For example, in three dimensions, 
I = [1 0 0 ; 0 1 0 ; 0 0 1].

Here is a column vector: [0 1 0]'.


x \mapsto y_1\ldots y_n


1/3


1 + 1 = 2
1 \leq 3 - 1
\leq 5


\ begin{graph}
    rankdir=LR
    v_1
    v_1 -> v_2 
    v_2 -> v_3 
    v_3 -> v_1 
\ end{graph}
\end{verbatim}

\section*{ download it today! at http://easylatex.sf.net }
\end{document}